\pdfoutput=1

\documentclass[11pt]{article}

\usepackage[preprint]{acl}

\usepackage{times}

\usepackage{enumitem}
\usepackage{tcolorbox}

\usepackage{latexsym}
\usepackage{algorithm} 
\usepackage{algpseudocode} 
\usepackage{amsmath} 
\usepackage{amssymb}
\usepackage{pifont}

\usepackage{amssymb}
\usepackage{xcolor}

\usepackage{graphicx}

\usepackage[T1]{fontenc}

\usepackage[utf8]{inputenc}

\usepackage{microtype}

\usepackage{inconsolata}

\usepackage{graphicx}
\usepackage{multirow}
\usepackage{amssymb}

\usepackage{soul}

\title{Beyond Fine-Tuning: In-Context Learning and Chain-of-Thought for Reasoned Distractor Generation}

\author{{\bf Elaf Alhazmi}$^{1,}$$^3$, {\bf Quan Z. Sheng}$^1$, {\bf Wei Emma Zhang}$^2$\\[1mm]
      $^1$School of Computing, Macquarie University, Australia\\
      $^2$School of Computer and Mathematical Sciences, Adelaide University, Australia\\
      $^3$College of Engineering and Computing in Al-Lith, Umm Al-Qura University, Saudi Arabia\\
        \texttt{elaf.alhazmi@hdr.mq.edu.au},  \texttt{eafhazmi@uqu.edu.sa},\\
        \texttt{michael.sheng@mq.edu.au}, 
        \texttt{wei.e.zhang@adelaide.edu.au}
       } 

\begin{document}
\maketitle
\begin{abstract} 
Distractor generation (DG) remains a labor-intensive task that still significantly depends on domain experts. The task focuses on generating plausible yet incorrect options, known as distractors, for multiple-choice questions. A reliable distractor must be contextually relevant to the question and able to mislead examinees through implicit reasoning when identifying the correct answer. While a recent method integrates fine-tuning pre-trained encoder-decoder models with contrastive learning to generate semantically relevant distractors for a given question-answer, it often fails to capture the underlying reasoning process that experts utilize when selecting distractors in benchmarks.  In this paper, we explore large language models (LLMs) reasoning for DG through in-context learning with unsupervised semantic retrieval for selecting few-shot examples. We design a rationale-augmented DG framework that jointly generates distractors and their rationales for a given question-answer. Extensive experiments on six benchmarks, with varying average distractor lengths and domains, demonstrate that prompting LLMs with few-shot examples substantially improves the performance compared to recent DG models. It outperforms recent approaches and achieves state-of-the-art results in generating reasoned distractors that align with human-labeled benchmarks.
\end{abstract}

\begin{figure}[t]
\begin{center}
    \includegraphics[width=0.50\textwidth, height = 6cm]{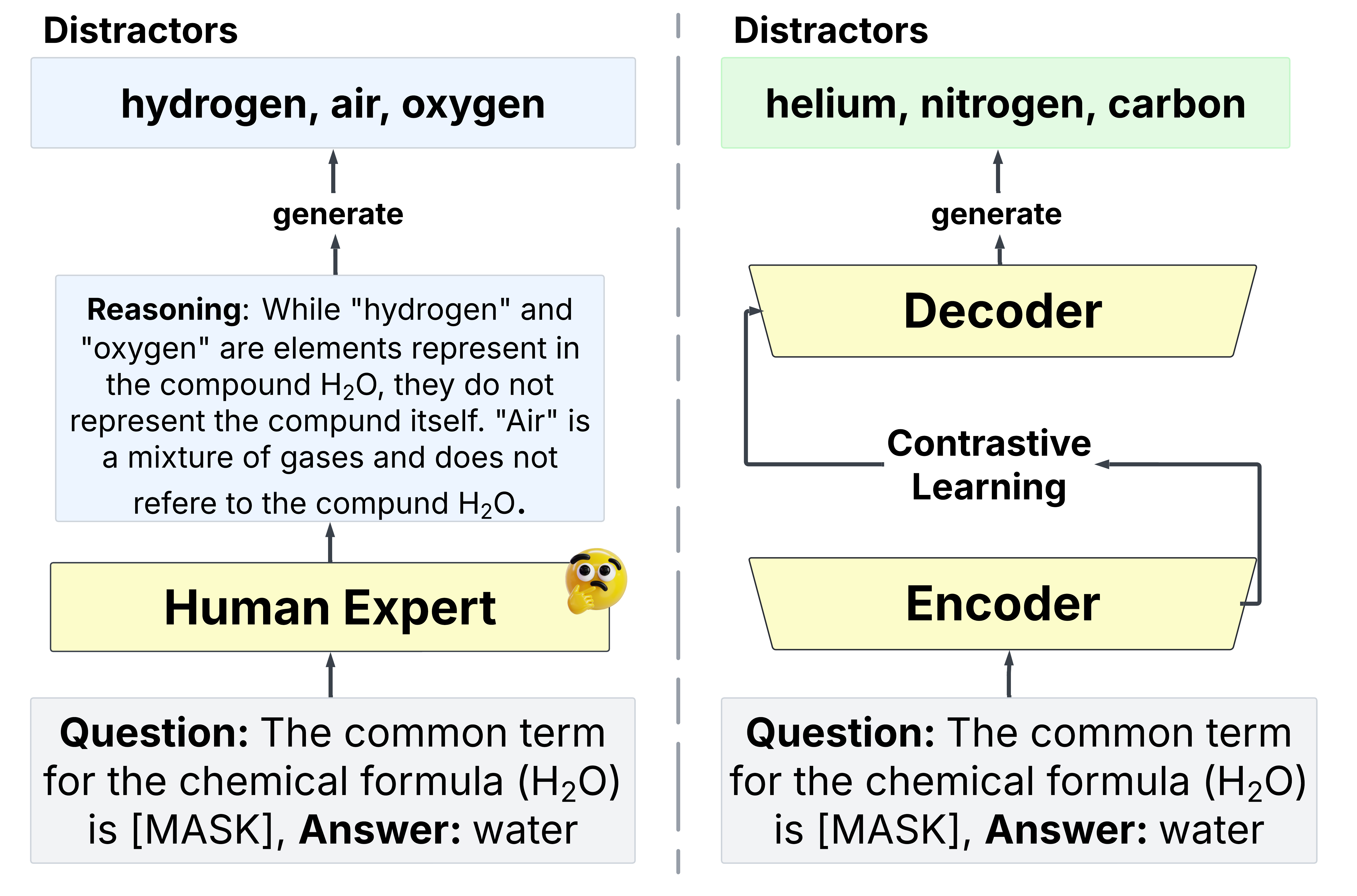}
 \vspace{-8mm}
     \caption{Generated distractors by human reasoning and contrastive-based pre-trained encoder-decoder models.}
    \label{fig:framework}
    \vspace{-8mm}
\end{center}
\end{figure}

\section{Introduction}
Distractor generation (DG) refers to the automated generation of misleading options in multiple-choice questions (MCQs) \cite{chen2022survey, alhazmi-etal-2024-distractor}. In practical assessment, the presence of well-selected {\em distractors} ensures that examinees can thoughtfully distinguish the correct answer from the wrong options. Unfortunately, selecting such distractors remains a challenging, labor-intensive, and expert-dominant task \cite{ch2018automatic, kurdi2020systematic, das2021automatic}.

The DG field has progressed considerably in methodologies, especially with transformer-based models, in textual \cite{xie-etal-2018-large}, mathematical \cite{feng-etal-2024-exploring}, and multi-modal \cite{yagcioglu-etal-2018-recipeqa} aspects. In text-based DG, an early work \cite{ren2021knowledge} proposed a {\em candidate generation and selection framework} that generates a set of distractors and selects the top candidates based on embedding models. Then, several works explored DG as a sequence-to-sequence (Seq2Seq) task, known as {\em Text2Text} architecture \cite {wang-etal-2023-distractor}, by fine-tuning pre-trained encoder-decoder models without a selection method.

Although \citet{alhazmi-etal-2025-fine} integrates contrastive learning as semantic representation learning in Text2Text models to emphasize in-context DG, they still struggle to align the reasoning behind generated distractors with human-labeled distractors in benchmarks, as indicated in Figure~\ref{fig:framework}.  For a question (the chemical formula for $H_{2}O$ is ...), a recent contrastive approach may generate valid distractors (carbon, nitrogen, helium) that are in-context, but they are often misaligned with human-label reasoning, e.g., ($H_{2}O$ is a compound made of two hydrogen atoms and one oxygen atom). Thus, \textit{hydrogen} and \textit{oxygen} may refer to more suitable pedagogically meaningful distractors.  In practice, experts rarely select distractors in isolation; they rely on explicit reasoning or rationale (i.e., step-by-step justification) to distinguish the answer from incorrect options. Reasoning-aware distractors play a critical role in effectively challenging examinees. The recent method \cite{alhazmi-etal-2025-fine} utilizes contrastive learning to improve semantic plausibility, but they fail to model such reasoning in DG.

The emerging LLMs \cite{NEURIPS2020_1457c0d6} have transformed downstream NLP tasks with {\em in-context learning} (ICL) \cite{liu2023pre, dong-etal-2024-survey}.  ICL allows LLMs to learn from a limited number of few-shot examples with a given input (\(x\)) and output (\(y\)) to generate a desired target for a given test input without fine-tuning or parameter updates. ICL is remarkably effective for a wide range of reasoning tasks, including mathematical problem solving \cite{NEURIPS2020_1457c0d6, NEURIPS2022_9d560961}, commonsense reasoning \cite{NEURIPS2022_8bb0d291, zhang-etal-2024-improving-diversity}, and question answering \cite{li-etal-2023-shot, patidar-etal-2024-shot}.

While recent works utilize ICL in math DG \cite{mcnichols2023exploring, feng-etal-2024-exploring} and multilingual DG benchmark \cite{bitew2023distractor}, the state-of-the-art (SOTA) text-based DG works still rely on supervised fine-tuning models \cite{chiang-etal-2022-cdgp, wang-etal-2023-distractor,taslimipoor-etal-2024-distractor-generation, yu-etal-2024-enhancing, alhazmi-etal-2025-fine}.

ICL framework is sensitive to example selection \cite{zebaze-etal-2025-context} when using random or 
semantic-embedding retrieval. While conventional DG methods formulated the task as mapping a given question-answer to a set of distractors, recent work has also shown that rationales, or step-by-step chain-of-thought (COT) \cite{NEURIPS2022_9d560961}, can improve the performance in multi-step reasoning tasks \cite{NEURIPS2022_8bb0d291}. We utilize COT to generate rationales for each question-answer and distractors in the training set and integrate ICL to generate distractors and rationales for a given question-answer test input. We demonstrate that integrating LLM reasoning to generate distractors can achieve more accurate results than conventional DG approaches.

The main contributions are (i) exploring ICL and the rationale-augmented DG framework for generating reasoned distractors; (ii) benchmarking our approach with SOTA DG models using both automatic and manual evaluation metrics; and (iii) achieving SOTA results in multiple benchmarks, including word-level and long-level distractors in science, general knowledge, and medical domains.


 \begin{figure*}[!tb]
\begin{center}
    \includegraphics[width=1\textwidth]{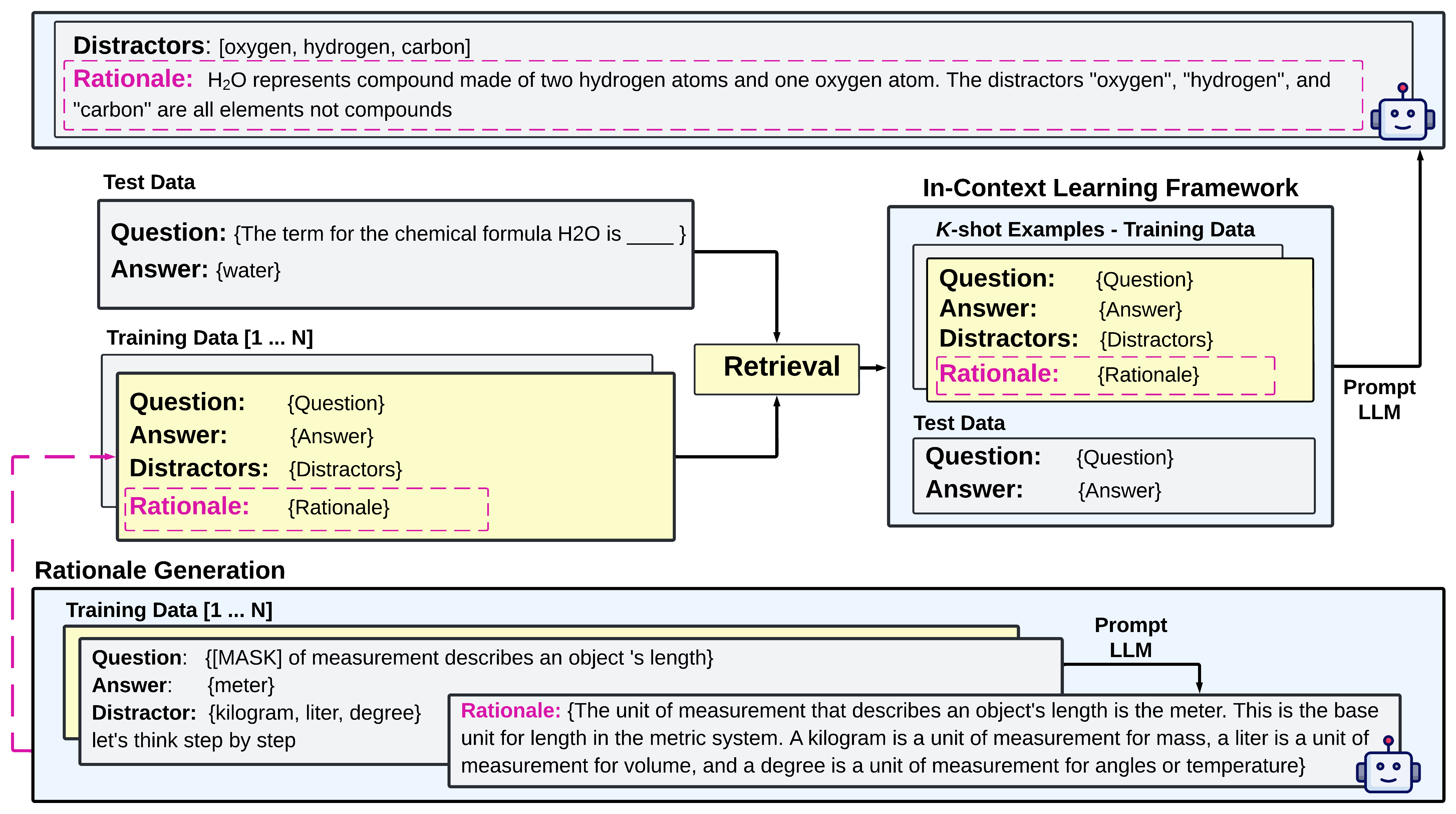}
        \vspace{-5mm}
    \caption{The in-context learning framework by large language model and chain-of-thought rationale generation for the training dataset. The dashed arrows indicate the optional augmented-rationale in-context learning framework.} 
    \label{fig:pipline_loss}
    \vspace{-5mm}
\end{center}
\end{figure*}

\section{Related Works} \label{sec:related_work}

\subsection{Distractor Generation (DG)}
Transformer-based pre-trained language models (PLMs) \cite{vaswani2017attention} have significantly advanced DG methods across multi-modal \cite{luo-etal-2024-chain}, mathematical \cite{fernandez2024divert}, and text-based \cite{qu-etal-2024-unsupervised} applications, including question answering \cite{liang-etal-2018-distractor} and reading comprehension \cite{maurya2020learning}.

Text-based approaches in MCQs include two main directions: {\em candidate generation and selection framework} (CSG-DS) \cite{ren2021knowledge} and {\em Text2Text architecture} \cite{wang-etal-2023-distractor}. The former utilizes a candidate generation step before selecting distractors with embedding models \cite{taslimipoor-etal-2024-distractor-generation}, while the latter utilizes fine-tuning pre-trained encoder-decoder models to generate distractors without a selection method.

Recently, the Text2Text DG approach \cite{yu-etal-2024-enhancing} has shown stronger performance compared to CSG-DS frameworks \cite{chiang-etal-2022-cdgp}. \citet{alhazmi-etal-2025-fine} introduced a contrastive-learning objective into Text2Text models to enhance the embedding representation learning between input and output in pre-trained encoder-decoder models to generate in-context distractors.

Despite these advances, the generated distractors often remain misaligned with human-labeled distractors. Human experts usually apply reasoning when selecting distractors for a given question and answer, a process not fully captured by recent fine-tuned pre-trained encoder-decoder models. ICL has emerged as a significant paradigm that leverages LLMs reasoning to generate distractors by prompting the model in zero-shot or few-shot learning. While prompting paradigms have been explored in text-based multilingual \cite{bitew2023distractor} and mathematical DG tasks \cite{feng-etal-2024-exploring}, fine-tuning PLMs remains more common in text-based DG benchmarks \cite{wang-etal-2023-distractor,taslimipoor-etal-2024-distractor-generation,yu-etal-2024-enhancing,alhazmi-etal-2025-fine}.

\subsection{In-Context Learning (ICL)}
LLMs demonstrate strong ICL \cite{dong-etal-2024-survey} capabilities, enabling them to solve unseen tasks by conditioning on a few-shot examples without parameter updates. ICL shows significant improvements in mathematical problem solving \cite{NEURIPS2020_1457c0d6, NEURIPS2022_9d560961}, commonsense reasoning \cite{NEURIPS2022_8bb0d291, zhang-etal-2024-improving-diversity}, and question answering \cite{li-etal-2023-shot} tasks.

Initially, ICL performance is highly sensitive to the selection of examples. \citet{liu-etal-2022-makes} proposed \textit{k}-nearest-neighbor (k-NN) unsupervised retrieval for selecting examples, while \citet{rubin-etal-2022-learning} introduced a two-stage framework that trains a dense retrieval to identify optimal examples. Similarly in DG domains, while \citet{feng-etal-2024-exploring} applied k-NN retrieval for mathematical DG, \citet{bitew2023distractor} trained supervised bi-encoder retrieval \cite{bitew2022learning} for multilingual DG. We extend the use of unsupervised k-NN retrieval for selecting in-context examples to text-based DG benchmarks.

Chain-of-thought (COT) \cite{NEURIPS2022_9d560961} encourages LLMs to generate explicit, step-by-step reasoning traces, known as rationales \cite{ling-etal-2017-program}, before predicting a final answer. \citet{NEURIPS2022_8bb0d291} outlined strong performance of zero-shot COT by prompting “Let’s think step by step". Integrating rationales can serve as auxiliary signals for improving ICL \cite{wang2022rationale, huang-etal-2023-large} or distilling reasoning \cite{tian2025beyond} into smaller pre-trained models.  We explore COT rationales as reasoning-based augmentation for DG.

\section{Methodology} \label{sec:methodology}
\vspace{-1mm}
Section~\ref{sec:task_formulation} formalizes the DG task, and Section~\ref{sec:ICL} introduces the ICL framework. The framework incorporates unsupervised similarity-based retrieval, described in Section~\ref{sec:unsupervised}, and COT rationale-based augmentation, presented in Section~\ref{sec:COT}.

\vspace{-1mm}
\subsection{Task Definition}  \label{sec:task_formulation}
\vspace{-1mm}
The DG task is defined as mapping (\({x} \mapsto {y}\)), where \( x \) denotes the input question \( \mathbf{Q} \) and its answer \( \mathbf{A} \), and \( y \) denotes the target sequence of distractors \( \mathbf{D} = \{d_{1}, d_{2}, \ldots, d_{N}\} \) with \( N > 0 \). This conditional text generation problem is defined as:
\begin{equation}
p_{{\phi}}(y \mid x) = \prod_{t=1}^{N} p_{\phi} (y_t \mid {y}_{<t}, {x})
\end{equation}
where \({\phi}\) are fixed large language model parameters, \( y_t \) represents the sequence of letters in the \( t \)-th distractor, and \({y}_{<t} \) denotes the sequences of all distractors generated before \( y_t \).

\subsection{In-Context Learning Framework} 
\label{sec:ICL}
ICL refers to the ability of large models to generate desired outputs from few-shot examples without updating model parameters. It extends conditional text generation using a context \({C}\) containing \({k}\) labeled examples, with the probability defined as:
\begin{equation}
p_{{\phi}}(y \mid C, x) = \prod_{t=1}^{N} p_{\phi} (y_t \mid {y}_{<t}, C, {x})
\end{equation}

Given a set of training example pairs
$e_{i} =\{(x_{\text{i}}, y_{\text{i}})\}_{i=1}^k$,
large language models predict distractors $\hat{y}$ based on the concatenation $\oplus$ of given examples  \((e_1 \oplus ...  \oplus e_k)\) as \(C\) and input $x_{test}$.
\begin{equation}
    \hat{y} = LLM (e_1 \oplus e_{2} \oplus ...  \oplus e_k\oplus x_{test})   
\end{equation}

\subsubsection{Unsupervised Retrieval} \label{sec:unsupervised}
To select few-shot examples from the training set as illustrated in Figure \ref{fig:pipline_loss}, we utilize a \textit{k}-nearest-neighbor (k-NN) retrieval \cite{liu-etal-2022-makes, feng-etal-2024-exploring}.  All training instances \(x_i\) are encoded using a sentence encoder $\mu_\theta(\cdot)$ to obtain their vector representations.  Given \(x_{\text{test}}\), we compute its embedding $v_{\text{test}}$ and identify the top-\(k\) most similar training instances using a similarity metric (e.g., cosine similarity), as outlined in Eq. \ref{eq:cosine_similarity}.  The retrieved neighbors \(x_1, \ldots, x_k\) are ordered by descending similarity, i.e., \(s(v_i, v_{\text{test}}) \ge s(v_j, v_{\text{test}})\) for \(i < j\). 
\begin{equation}
     \text{s}(\mathbf{v}_i, \mathbf{v}_j) = \frac{\mathbf{v}_i^\top \mathbf{v}_j}{\|\mathbf{v}_i\|_2 \|\mathbf{v}_j\|_2}
     \label{eq:cosine_similarity}
\end{equation}

Each retrieved \(x_i\) is concatenated with its corresponding target \(y_i\) to construct the context \(C\).

\subsubsection{Chain-of-Thought Augmentation} \label{sec:COT}
We introduce rationale augmentation using a zero-shot COT. We prompt a model to generate a rationale \( r \) for a given training question-answer \( x \) and its distractors \( y \) as illustrated in Figure \ref{fig:pipline_loss}.

Then, given a set of training examples $e_{r_i} = \{(x_i, y_i, r_i)\}_{i=1}^{k}$, the large language model predicts $\hat{z}$, consisting of distractors $\hat{y}$ and rationale $\hat{r}$, based on the concatenation $\oplus$ of given retrieved examples $(e_{r_1} \oplus \cdots \oplus e_{r_k})$ as \(C\) and input $x_{\text{test}}$.
\begin{equation}
    \hat{z} = LLM(e_{r_1} \oplus e_{r_2} \oplus \cdots \oplus e_{r_k} \oplus x_{test})
\end{equation}

Algorithm~\ref{alg:k-NN_retrieval} in Appendix~\ref{sec:k-NN_app} describes the k-NN-based retrieval and its optional COT augmentation.

\section{Experiments} 
\label{sec:experiment}

\subsection{Datasets} 
\label{sec:datasets}
We conduct the experiments on six datasets, including science (SciQ \cite{welbl-etal-2017-crowdsourcing}, MCQL \cite{liang-etal-2018-distractor}, ARC \cite{clark2018think}), general knowledge (MCQ \cite{ren2021knowledge}), and medical domain (MedQA \cite{jin2021disease}) as detailed in Appendix \ref{sec:datasets_domains}. Each dataset contains one answer and three to four distractors, with different token lengths. The statistics are outlined in Table~\ref{tab:Multiple Choice Datasets}.

\begin{table}[h]
\centering
\resizebox{0.88\columnwidth}{!}{%
\begin{tabular}{c|c|c|c|c}
\hline \hline
\bf Datasets & \bf Train & \bf Valid & \bf Test & \bf All \\ \hline

MCQ             & 1,856   & 465    & 259   & 2,580     \\ 
\hline
SciQ            & 11,700  & 1,000  & 1,000   & 13,700 \\
\hline
MCQL            & 6,362   & 600    & 600   & 7,562    \\
\hline
ARC-Easy        &  2,251  &  570   & 2,376  &  5,197  \\ 
\hline
ARC-Challenge   &  1,119  &  299   & 1,172  &  2,590   \\ 
\hline
MedQA           &  10,178  & 1,272  & 1,273  &  12,723 \\ 
\hline\hline
\end{tabular}%
}
\vspace{-2mm}
\caption{Statistics of the Datasets.}
\label{tab:Multiple Choice Datasets}
\vspace{-5mm}
\end{table}

\subsection{Baselines Models}
\vspace{-1mm}
\textbf{CSG-DS}: We fine-tune T5 \cite{raffel2020exploring} model as a candidate-generation to generate ten distractors; then we select the top three with beam search \cite{gao2019generating}.  
We also use T5 and GPT-3 \cite{NEURIPS2020_1457c0d6}, with few-shot k-NN retrieval for candidate generation, followed by a clustering selection method \cite{taslimipoor-etal-2024-distractor-generation}.

\vspace{1mm}
\noindent \textbf{Text2Text Architecture}:
We train T5 as Seq2Seq model. We explore multi-task learning \cite{wang-etal-2023-distractor} and contrastive learning \cite{alhazmi-etal-2025-fine} with InfoNCE \cite{oord2018representation} objective.

\vspace{1mm}
\noindent \textbf{Instruction-tuning}: We fine-tune TinyLlama \cite{zhang2024tinyllama} under an instruction-tuning paradigm to evaluate whether a lightweight 1.1B model can effectively learn the DG task.

\vspace{1mm}
\noindent \textbf{Prompting}: We utilize {zero-shot} prompting and {random} retrieval for ICL and COT augmentation.

\vspace{-1mm}
\subsection{Evaluation Metrics}
\vspace{-1mm}
We use ranking-based metrics \cite{wang-etal-2023-distractor} to assess how well models retrieve relevant distractors in the top-$k$ positions. {\em Order-unaware} metrics include F1 score (F1@K), precision (P@K), and recall (R@K), while {\em order-aware} metrics include mean reciprocal rank (MRR@K) and normalized discounted cumulative gain (NDCG@K). We report percentage-based indicators that quantify answer repetition ($I_a$) and distractor repetition ($I_d$).

We also utilize human evaluation metrics, including {\em relevance}, which measures if the distractors are in-context to a given input; {\em difficulty}, which captures the ability to mislead from the correct answer; and {\em fluency}, which ensures the distractors are valid and not duplicated.  We select five questions per dataset and six evaluators: three academics with over two years of experience to assess 25 science and general-knowledge questions and three medical doctors with over two years of clinical experience to evaluate the remaining five medical questions, using a five-point rating system from 1 to 5. We measure {\em accuracy} using question-answering models with generated distractors.

\begin{table*}[tb!]
\centering
\resizebox{0.93\textwidth}{!}{
\setlength{\tabcolsep}{6pt}
\begin{tabular}{c|c|l|c|c|c|c|c|c c}
\hline\hline
\textbf{Dataset} & \textbf{Approach} & \multicolumn{1}{c|}{\textbf{Model}}
& \textbf{P@1} & \textbf{R@1} & \textbf{F1@3} & \textbf{MRR} & \textbf{NDCG@3} & $\mathbf{I_a}$ (\%) & $\mathbf{I_d}$ (\%) \\
\hline
\multirow{12}{*}{\textbf{MCQ}}
& \multirow{3}{*}{\bf CSG-DS}
& T5-CSG-DS(beam)
& 17.37 & 5.79 & 13.38   & 24.20    & 30.11 &  36.29 &  0.39  \\
& &T5-CSG-DS(clustering)         
& 11.58  &  3.86 &  7.72  &   16.47   &  20.96   &  18.15   &   0.0      \\
& & GPT-3-CSG-DS(clustering)         
&  14.29   & 4.76 & 8.49  &  18.92   &  22.95   &   0.39   &   0.00  \\
\cline{2-3}
& \multirow{3}{*}{\bf Text2Text}
& T5(base)   
& 14.29 & 4.76  & 10.81	& 20.14 & 24.68   &  17.76  & 18.15 \\
& & T5(multi-task) 
& 15.06 & 5.02 &  12.23  &  19.95 &  23.54 &  19.31 &   15.83  \\
& & T5(contrast)
& 22.78 & 7.59  & 15.70  & 28.57  & 32.33  & 14.67  &  10.81   \\
\cline{2-3}
& \multirow{1}{*}{\bf Instruction Tuning}
& TinyLlama  
&  22.78    &   7.59 &  15.06    &   28.44  &   33.16     &   3.09  & 1.54   \\
\cline{2-3}
& \multirow{5}{*}{\bf Prompting}
& GPT-3(zero-shot)   
& 25.87 & 8.62 & 17.63 &  32.56 & 38.33  &  0.00  &  0.00   \\
& & GPT-3(few-shot-random)  
& 25.87 & 8.62 & 18.28  &  33.72   &  40.68   &   0.00   &  0.00 \\
& & GPT-3(few-shot-k-NN)   
& 29.73 & 9.91 &  {\bf 19.69}  & {\bf 36.62} & {\bf 42.37}&  0.00   &  0.00    \\
& & GPT-3(few-shot-COT-random) 
& 28.96 & 9.65 & 19.05 & 36.04 & 42.08 &    0.00   &  0.00    \\
& & GPT-3(few-shot-COT-k-NN)
& {\bf 30.50}  & {\bf 10.17} & 19.43 & 36.29 & 40.60 &  0.00   & 0.00    \\
\hline

\multirow{12}{*}{\textbf{SciQ}}

& \multirow{3}{*}{\bf CSG-DS}
& T5-CSG-DS(beam) 
&20.30 & 6.77 & 14.33 & 26.20 & 31.30 &  29.10 & 0.20  \\
& &T5-CSG-DS(clustering)         
&  10.50 & 3.50 & 6.30   &   14.35   & 17.95    &   22.3     &   0.0      \\
& & GPT-3-CSG-DS(clustering)         
& 10.80 & 3.60 & 5.20 & 13.33 & 15.73 &  0.30 &  0.00 \\
\cline{2-3}
& \multirow{3}{*}{\bf Text2Text}
& T5(base)     
& 18.90 & 6.30 & 13.77 & 23.23  & 26.66  &  23.30 &  12.80 \\
& & T5(multi-task)  
& 21.20 & 7.07 & 14.27 & 25.47 & 28.52 &  18.90 & 11.10  \\
& & T5(contrast)  
& 25.00 & 8.33 & {\bf 17.73} & {\bf 31.42} & {\bf 36.68}  & 8.80 &  6.50 \\
\cline{2-3}
& \multirow{1}{*}{\bf Instruction Tuning}
& TinyLlama  &  20.50     & 6.83   &  15.40    & 27.07    & 32.73 & 2.40 & 1.90        \\
\cline{2-3}
& \multirow{5}{*}{\bf Prompting}
& GPT-3(zero-shot)  
& 24.40 & 8.13 & 15.47 & 29.30  & 33.45 & 0.10 & 0.30 \\
& & GPT-3(few-shot-random)  
& 25.10 & 8.37 & 16.13 & 30.17 & 34.57  &  0.20 & 0.10 \\
& & GPT-3(few-shot-k-NN) 
& 24.90 & 8.30  &16.53  & 30.45  &35.24  & 0.10 &  0.00 \\
& & GPT-3(few-shot-COT-random) 
& 24.50 & 8.17  & 15.70 & 29.53  & 33.82 & 0.20  & 0.00  \\
& & GPT-3(few-shot-COT-k-NN) 
& {\bf 25.50}  & {\bf 8.50} & 17.03 & 31.08 & 35.89  & 0.10 &  0.00 \\
\hline

\multirow{12}{*}{\textbf{MCQL}}
& \multirow{3}{*}{\bf CSG-DS}
& T5-CSG-DS(beam)  
&20.33 & 6.78  & 14.50 & 25.42 & 29.65 & 37.17 & 1.83 \\
& &T5-CSG-DS(clustering)         
& 12.17  & 4.06 &  5.56  &  15.69    &  19.04    &   20.5   &   0.00      \\
& & GPT-3-CSG-DS(clustering)         
& 11.17 & 3.72 & 4.28 & 13.92 & 16.55 & 0.17 & 0.00 \\
\cline{2-3}
& \multirow{3}{*}{\bf Text2Text}
& T5(base)   
&17.83  & 5.94 & 10.78 & 20.07 & 21.68 & 28.83 & 4.83 \\
& & T5(multi-task)  
& 19.33 & 6.44 & 11.50 & 22.14 & 24.45  & 21.5 & 4.17 \\
& & T5(contrast)   
& 23.00 & 7.67 & 13.67 & 26.42 & 29.13 & 16.67 &  3.67 \\
\cline{2-3}
& \multirow{1}{*}{\bf Instruction Tuning}
& TinyLlama  &   28.83    & 9.61   &20.56 &   33.33  &   37.03 & 2.67 & 1.17   \\
\cline{2-3}
& \multirow{5}{*}{\bf Prompting}
& GPT-3(zero-shot)   
& 24.17 & 8.06 & 13.78 & 27.03 & 29.37 & 0.00 & 0.17  \\
& & GPT-3(few-shot-random)  
& 30.17 & 10.06 & 19.06	 & 34.36 & 37.98 & 0.00  & 0.50  \\
& & GPT-3(few-shot-k-NN)   
& 35.33 & 11.78 & {\bf 24.44}  & 39.61 & 42.81 & 0.00 & 0.33  \\
& & GPT-3(few-shot-COT-random)   
& 32.00 & 10.67 & 19.44 & 35.58 & 38.51 & 0.00 &  0.17 \\
& & GPT-3(few-shot-COT-k-NN) 
& {\bf 36.17}  & {\bf 12.06}  & 23.61 & {\bf 39.94} & {\bf 42.90} & 0.00  & 0.00  \\
\hline

\multirow{12}{*}{\textbf{ARC-Easy}}
& \multirow{3}{*}{\bf CSG-DS}
& T5-CSG-DS(beam)  
& 12.75 & 4.28 & 9.56 &  16.88 & 20.45 & 33.63 & 0.63 \\
& &T5-CSG-DS(clustering)         
&  7.95 &  2.68   &  4.50  &  10.67    &  13.16   &   16.16     &   0.00      \\
& & GPT-3-CSG-DS(clustering)         
& 9.81 & 3.28 & 5.55 & 13.06 & 16.08 &  0.08 &  0.00 \\
\cline{2-3}
& \multirow{3}{*}{\bf Text2Text}
& T5(base) 
& 11.24 & 3.77 & 7.38 & 14.05 & 16.30 & 13.97 & 7.70 \\
& & T5(multi-task) 
& 10.06 & 3.38 & 6.97  & 12.73 & 14.9 & 13.51 & 5.77 \\
& & T5(contrast) 
& 13.85 & 4.64 & 10.39 & 17.53 & 20.45 & 6.69 & 3.41 \\
\cline{2-3}
& \multirow{1}{*}{\bf Instruction Tuning}
& TinyLlama  &   15.03    &  5.03  &   12.73   &  19.76   & 23.66 &  1.98 & 1.05   \\
\cline{2-3}
& \multirow{5}{*}{\bf Prompting}
& GPT-3(zero-shot)   
& 20.75 & 6.94  & 15.28 & 25.55 & 29.32 &  0.00  &  0.00  \\
& & GPT-3(few-shot-random) 
& 23.19 & 7.75 & 17.74 & 28.37  & 32.41 & 0.00 & 0.04  \\
& & GPT-3(few-shot-k-NN)
& {\bf 24.24} & {\bf 8.10} & {\bf 18.10} & {\bf 29.08} & {\bf 32.82} & 0.00   & 0.17  \\
& & GPT-3(few-shot-COT-random) 
& 23.70 & 7.92 & 17.77 & 28.67 &32.44  & 0.00   &  0.00  \\
& & GPT-3(few-shot-COT-k-NN)
& 23.70 & 7.92 & 17.95 & 28.36 & 31.74 &  0.00  &  0.00  \\
\hline

\multirow{12}{*}{\textbf{ARC-Challenge}}

& \multirow{3}{*}{\bf CSG-DS}
& T5-CSG-DS(beam)  
& 7.85  & 2.67 & 6.50 & 11.46 & 14.71 & 31.83 & 0.26 \\
& &T5-CSG-DS(clustering)         
&  5.46 &   1.86  &   2.23   &  7.35   &   9.18     &   12.97   &  0.00  \\
& & GPT-3-CSG-DS(clustering)         
& 8.70 & 2.92 & 4.00 &  10.75 & 12.68 & 0.00  & 0.00 \\
\cline{2-3}
& \multirow{3}{*}{\bf Text2Text}
& T5(base)     
& 7.17 & 2.45 & 3.68 & 8.79 & 10.14 & 19.20 & 2.90 \\
& & T5(multi-task)         
& 5.38 & 1.85 & 3.63 & 7.21  & 8.81 &  18.43 & 2.39 \\
& & T5(contrast)  
& 9.13 & 3.09 & 5.47 & 11.26 & 13.06 & 9.64 & 3.07 \\
\cline{2-3}

& \multirow{1}{*}{\bf Instruction Tuning}
& TinyLlama  &   12.37 &   4.15 &   8.60   &  15.15   & 17.55  &  1.37  & 1.11 \\
\cline{2-3}
& \multirow{5}{*}{\bf Prompting}
& GPT-3(zero-shot)   
& 17.41  & 5.85 & 11.96 & 20.31	 & 22.53 & 0.09 & 0.09 \\
& & GPT-3(few-shot-random)  
& {\bf 21.08} & {\bf 7.07} & 14.96 & {\bf 24.82} & {\bf 27.75} & 0.26  & 0.17 \\
& & GPT-3(few-shot-k-NN)   
& 20.48 & 6.87 & 15.05 & 24.06 & 26.80 & 0.17  & 0.09  \\
& & GPT-3(few-shot-COT-random)   
& 20.31 & 6.81 & {\bf 15.29} & 24.23 & 27.23 &  0.34 & 0.09 \\
& & GPT-3(few-shot-COT-k-NN)  
& 19.88 & 6.67 & 14.60 & 23.25 & 25.73  & 0.17  & 0.09 \\
\hline

\multirow{12}{*}{\textbf{MedQA}}
& \multirow{3}{*}{\bf CSG-DS}
& T5-CSG-DS(beam) 
& 6.83 & 1.71 & 4.44 & 9.54 & 12.00 & 21.05 & 0.55 \\
& &T5-CSG-DS(clustring)         
&  4.24 & 1.06 & 1.89   & 5.85  & 7.38    &    10.05    &  0.00       \\
& & GPT-3-CSG-DS(clustring)         
& 9.82 & 2.45 & 5.12 & 12.80 & 15.47 & 0.08 &  0.00 \\
\cline{2-3}
& \multirow{3}{*}{\bf Text2Text}
& T5(base)   
& 5.42 & 1.36 & 2.51 & 6.39 & 7.10 & 13.28 &  0.08 \\
& & T5(multi-task) 
& 4.95  & 1.24 & 2.60 & 6.11 & 7.00 & 11.31  &  0.08 \\
& & T5(contrast)  
& 6.83 & 1.71 & 3.50 & 8.42 & 9.57 & 9.19 & 0.31 \\
\cline{2-3}
& \multirow{1}{*}{\bf Instruction Tuning}
& TinyLlama  &   11.86 &  2.97  &  7.90    &  15.95  &  19.60  &  1.02  & 0.55       \\
\cline{2-3}
& \multirow{5}{*}{\bf Prompting}
& GPT-3(zero-shot)  
& 12.49 & 3.12 & 7.77 & 16.17 & 19.42 & 0.00 &  0.00 \\
& & GPT-3(few-shot-random)
& 17.44 & 4.36 &  11.90 & 23.10  & 26.50 & 0.00  & 0.00  \\
& & GPT-3(few-shot-k-NN) 
& 20.90 & 5.22 & {\bf 15.08} & {\bf 27.44} & {\bf 30.52} & 0.00 &  0.08 \\
& & GPT-3(few-shot-COT-random) 
& 17.20 & 4.30 & 11.74  & 22.52	 &  25.63 &  0.00 &  0.00  \\
& & GPT-3(few-shot-COT-k-NN)  
& {\bf 21.05} & {\bf 5.26} & 14.66 & 27.39 & 30.34 & 0.00  & 0.00   \\
\hline\hline
\end{tabular}
    }
\vspace{-2mm}
\caption{
Automatic evaluation results. The standard in-context learning framework is denoted as (few-shot), while chain-of-thought augmentation is referred to as (few-shot-COT). The best ranking-based scores are outlined in bold.
}
\label{tab:results_main}
\end{table*}

\begin{figure*}[!t]
    \centering
    \begin{minipage}{0.48\textwidth}
        \centering
        \includegraphics[width=\linewidth]{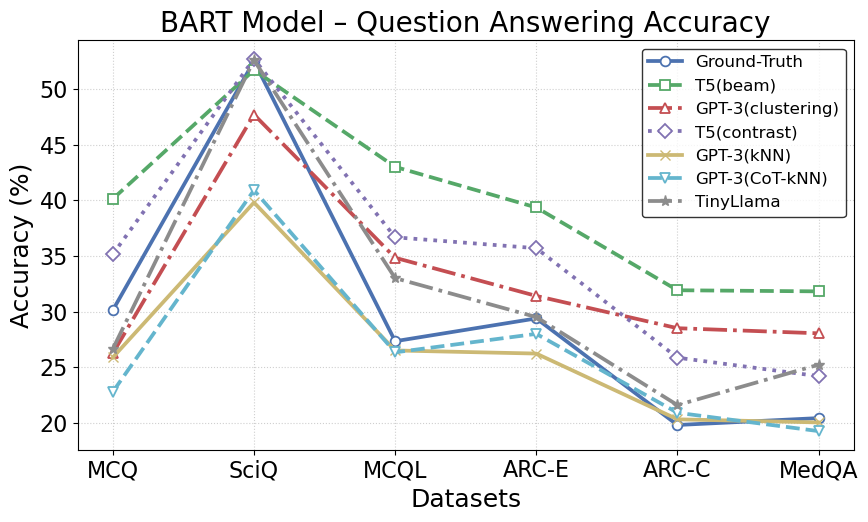}
        \vspace{-7mm}
        \caption{BART – Question Answering Accuracy}
        \label{fig:bart_accuracy}
    \end{minipage}
    \hfill
    \begin{minipage}{0.48\textwidth}
        \centering
        \includegraphics[width=\linewidth]{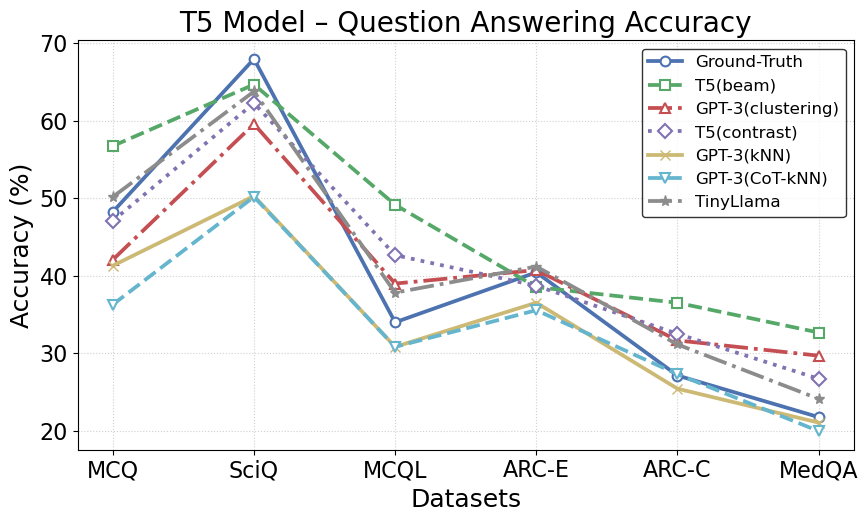}
        \vspace{-7mm}
        \caption{T5 – Question Answering Accuracy}
        \label{fig:t5_accuracy}
    \end{minipage}
    \vspace{-3mm}
\end{figure*}


\begin{table}[tb!]
\centering
\resizebox{0.9\columnwidth}{!}{
\begin{tabular}{l|c|c|c}
\hline\hline
\textbf{Model} & \textbf{Relevance} & \textbf{Difficulty} & \textbf{Fluency} \\ 
\hline
GPT-3 (COT-k-NN) 
& 3.74 & 3.15 & 4.65 \\ 

GPT-3 (k-NN) 
& \textbf{4.14} & \textbf{3.51} & \textbf{4.78} \\ 

TinyLlama 
& 3.34 & 2.97 & 4.47 \\ 

GPT-3 (clustering) 
& 3.45 & 2.93 & 4.56 \\ 

T5 (beam) 
& 3.00 & 2.51 & 3.90 \\ 

T5 (contrast) 
& 2.54 & 2.18 & 3.58 \\ 

Ground-truth  
& 3.73 & 3.28 & 4.58 \\    

\hline\hline
\end{tabular}
}
\vspace{-2mm}
\caption{Human evaluation results.}
\label{tab:human-evaluation}
\vspace{-5mm}
\end{table}

\vspace{-1mm}
\subsection{Implementation Details} 
\vspace{-1mm}
We use GPT-3.5-turbo \cite{NEURIPS2020_1457c0d6} for ICL with \(k = 5\), MPNet sentence encoder \cite{reimers2019sentence} for vector representations, and cosine similarity as base similarity metric.
Prompt templates and all model configurations are detailed in Appendix~\ref{sec:templates} and Appendix~\ref{sec:models}, 
respectively\footnote{\url{https://github.com/reasoning-dis/icl-COT}}.

\begin{table*}[tb!]
\centering
\resizebox{0.80\textwidth}{!}{
\setlength{\tabcolsep}{6pt}
\begin{tabular}{c|l|c|c|c|c|c|c|c}
\hline\hline
\textbf{Dataset} & \multicolumn{1}{c|}{\textbf{Model}}
& \textbf{P@1} & \textbf{R@1} & \textbf{F1@3} & \textbf{MRR} & \textbf{NDCG@3} & $\mathbf{I_a}$ (\%) & $\mathbf{I_d}$ (\%) \\
\hline
\multirow{6}{*}{\textbf{MCQ}}
& GPT-3(k-NN)       
& 29.73  & 9.91  & 19.69  & 36.62  &  42.37 &  0.00 & 0.00  \\
& GPT-3(k-NN)/wo-a   
& 24.32 &	8.11 &	15.57&	29.99&	34.71 &  0.39 & 1.16\\
& GPT-3(k-NN)/MiniLM
& 29.73 	& 9.91	&19.82	&35.78	& 40.42  &  0.00 & 0.77\\
& Mistral(k-NN)    
& 27.41	 &9.14	& 18.79  	& 34.3	 & 40.15 & 0.00 & 0.00 \\
& GPT-3(k-NN)/k-30       
& {\bf 30.12}	& {\bf 10.04}	& {\bf 21.62}	& {\bf 38.42}	& {\bf 45.29} & 0.00 & 0.39 \\
\hline
\multirow{6}{*}{\textbf{SciQ}}
& GPT-3(k-NN)      
& 24.90	& 8.30	& 16.53	 & 30.45	 & 35.24  & 0.10 & 0.00 \\
& GPT-3(k-NN)/wo-a  
& 19.50	& 6.50	& 13.47	& 24.33	& 28.46 &  0.30 & 0.80  \\
& GPT-3(k-NN)/MiniLM
& 24.90	& 8.30	& 16.77	& 30.75	& 35.87 & 0.00 & 0.10 \\
& Mistral(k-NN)     
& 20.10	& 6.70	& 15.50	 & 26.80  & 32.62  & 0.10 & 0.00  \\
& GPT-3(k-NN)/k-30      
& {\bf 25.90}	 & {\bf 8.63}	 & {\bf 18.10}	& {\bf 32.18}	 & {\bf 37.61}  & 0.10 & 0.10 \\
\hline
\multirow{6}{*}{\textbf{MCQL}}
& GPT-3(k-NN)        
& 35.33	 & 11.78	& 24.44	 &39.61	 & 42.81  &  0.00 & 0.33 \\
& GPT-3(k-NN)/wo-a   
& 23.50	&7.83	&16.56	&27.25	&30.33 & 1.17 & 3.33 \\
& GPT-3(k-NN)/MiniLM
& 37.50	& 12.50	& 24.28	& 40.92	& 43.40 & 0.00 & 0.17 \\
& Mistral(k-NN)       
& 27.67	&9.22	&18.89	&32.14	&35.93  & 0.00 & 0.17 \\
& GPT-3(k-NN)/k-30
& {\bf 38.17}	 & {\bf 12.72}	 & {\bf 26.06}	 & {\bf 42.17}	 & {\bf 45.19}  & 0.17 & 0.33 \\
\hline
\multirow{6}{*}{\textbf{ARC-Easy}}
& GPT-3(k-NN)        
&24.24	&8.10	&18.10	&29.08	& 32.82   & 0.00 & 0.17 \\
& GPT-3(k-NN)/wo-a      
&15.45	&5.16	&12.01	&18.82	&21.41 & 0.13 & 0.84  \\
& GPT-3(k-NN)/MiniLM
&23.74	&7.93	&17.95	&28.70	&32.51  & 0.00 & 0.04\\
& Mistral(k-NN)       
& 17.93	 & 6.00	 & 14.04  & 22.46	 & 26.14  &  0.17 & 0.04 \\
& GPT-3(k-NN)/k-30      
& {\bf 25.93}	& {\bf 8.66}	& {\bf 19.38}	& {\bf 30.77}	& {\bf 34.37}  & 0.00 & 0.25 \\
\hline
\multirow{6}{*}{\textbf{ARC-Challenge}}
& GPT-3(k-NN)       
& 20.48	&6.87	&15.05	&24.06	&26.80  &  0.17 & 0.09\\
& GPT-3(k-NN)/wo-a       
& 13.91	&4.66	&10.46	&16.54	&18.48  &  0.51 & 1.11\\
& GPT-3(k-NN)/MiniLM
& 19.80	&6.64	&14.67	&23.69	&26.85 &  0.26 & 0.17\\
& Mistral(k-NN)         
& 14.59	&4.91	&10.26	&17.09	&19.20   & 0.68 & 0.17\\
& GPT-3(k-NN)/k-30      
& {\bf 21.25}	 & {\bf 7.13}	 & {\bf 16.01}	& {\bf 25.46}	  &  {\bf 28.77}  &  0.17 & 0.17 \\
\hline
\multirow{6}{*}{\textbf{MedQA}}
& GPT-3(k-NN)       
& 20.90	& 5.22	&15.08	&27.44	&30.52  &  0.00 & 0.08\\
& GPT-3(k-NN)/wo-a       
& 4.87	& 1.22	&3.61	&6.99	&8.66   & 0.00 & 0.08  \\
& GPT-3(k-NN)/MiniLM
& 19.80	& 4.95	& 13.85	& 25.65	& 28.36  & 0.08 & 0.08 \\
& Mistral(k-NN)      
& 14.61	& 3.65	& 10.35	&19.48	&23.46  & 0.94 & 0.00\\
& GPT-3(k-NN)/k-30       
&{\bf 22.00}	& {\bf 5.50}	& {\bf 15.42}	&{\bf 28.45}	& {\bf 31.52}  & 0.00 & 0.16\\
\hline\hline
\end{tabular}
}
\vspace{-2mm}
\caption{Ablation experimental results. wo-a means without answer and k-30 indicates number of examples.}
\label{tab:results_ablation}
\vspace{-3mm}
\end{table*}

\subsection{Evaluation Results}
\subsubsection{Automatic Evaluation Results}
Table \ref{tab:results_main} provides a summary of the automatic evaluation results across all datasets. ICL, with LLMs such as GPT-3, substantially improves DG performance compared to recent SOTA approaches.

From the table, we can observe that ICL significantly outperforms CSG-DS framework results across all datasets and metrics. It increases the beam-search candidate-generation F1@3 score from 13.38 to 19.69 on MCQ, 6.50 to 15.29 on ARC-Challenge, and 4.44 to 15.08 on MedQA. This result underscores the strength of ICL for DG with longer distractors and diverse domains.

ICL also outperforms Text2Text models. While the contrastive-based model provides steady improvements over base fine-tuning and multi-task learning in all datasets and still achieves the highest F1@3 score on SciQ, it remains less competitive with longer distractor tokens and diverse domains, such as the medical field. In fact, ICL shows consistently stronger performance than recent SOTA contrastive-based models in the DG task. Its F1@3 score rises from 15.70 to 19.69 on MCQ and 13.67 to 24.44 on MCQL, with even larger improvements on datasets with longer distractor lengths: 10.39 to 18.10 on ARC-Easy, 5.47 to 15.29 on ARC-Challenge, and 3.50 to 15.08 on MedQA.

Notably, ICL with LLMs provides substantial improvements over fine-tuning small language models (SLMs), including TinyLlama, which shows in some cases stronger automatic results than recent approaches (e.g., CSG-DS and Text2Text), particularly on datasets with longer distractor tokens. It raises the F1@3 score from 12.73 to 18.10 on ARC-Easy, 8.60 to 15.29 on ARC-Challenge, and 7.90 to 15.08 on MedQA. This highlights the reasoning ability of large models to address the DG task effectively (i.e., closer to human reasoning), without fine-tuning or any parameter updates.

ICL presents stronger results when combined with unsupervised k-NN retrieval for selecting few-shot examples. It significantly enhances the performance across several datasets. Compared to random retrieval, k-NN selection boosts the F1@3 score from 19.06 to 24.44 on MCQL and 11.90 to 15.08 on MedQA, while yielding modest improvements from 18.28 to 19.69 on MCQ, 16.13 to 16.53 on SciQ, 17.74 to 18.10 on ARC-Easy, and 14.96 to 15.05 on ARC-Challenge. This demonstrates that embedding semantically similar examples enables LLMs to generate distractors that closely align with human-labeled distractors and reasoning.

Although COT augmentation does not consistently improve all automatic metrics compared to base ICL, it achieves higher P@1 and R@1 on datasets such as MCQ, SciQ, MCQL, and MedQA. It also enhances the interpretability of the DG framework by revealing the reasoning behind predicted distractors and achieves the second-best performance in multiple datasets (e.g., MCQ, MCQL, ARC, and MedQA) compared to recent models.

Overall, ICL consistently outperforms zero-shot prompting and is far less prone to generating the correct answer or repeating distractors. CSG-DS with beam search shows the highest rates of answer leakage and distractor repetition, followed by standard Text2Text models. Although contrastive-based Text2Text model reduces these errors, both Text2Text and CSG-DS methods still fall short of the reliability (i.e., distractors match the correct answer) needed for high-quality DG.

\subsubsection{Human Evaluation Results}
Table \ref{tab:human-evaluation} reports human evaluation results for DG models. ICL with k-NN retrieval achieves the highest scores across all human metrics. COT augmentation exhibits a slightly lower difficulty score compared to ground truth, but its overall evaluation remains closely aligned with human-generated distractors. This indicates the effectiveness of LLMs reasoning in generating high-quality distractors.

We can also observe that recent approaches present a strong ability to generate relevant and fluent distractors, often comparable to the ground truth. Instruction-tuned models and GPT-3–based candidate generation achieve higher scores than contrastive Text2Text models and T5 beam search. This is largely because the latter models remain vulnerable to answer leakage or option repetitions, thereby reducing their quality during the evaluation. In fact, all approaches are able to generate sufficiently in-context distractors, making the human evaluation more subjective and less discriminative. 

We measure the accuracy for a given question-answer and generated distractors by each approach. As shown in Figure~\ref{fig:bart_accuracy} and Figure~\ref{fig:t5_accuracy}, ICL and its COT-augmented variants yield the lowest accuracy across all datasets in both BART and T5 models. Also, ground-truth distractors achieve low accuracy on several datasets. These results indicate that ICL yields distractors that more closely reflect human expert reasoning compared with recent DG approaches. Table \ref{tab:t5_bart_accuracy} in Appendix \ref{sec:models} provides the detailed accuracy results for both models.

\begin{figure}[t]
\begin{center}
    \includegraphics[width=0.48\textwidth, height = 4cm]{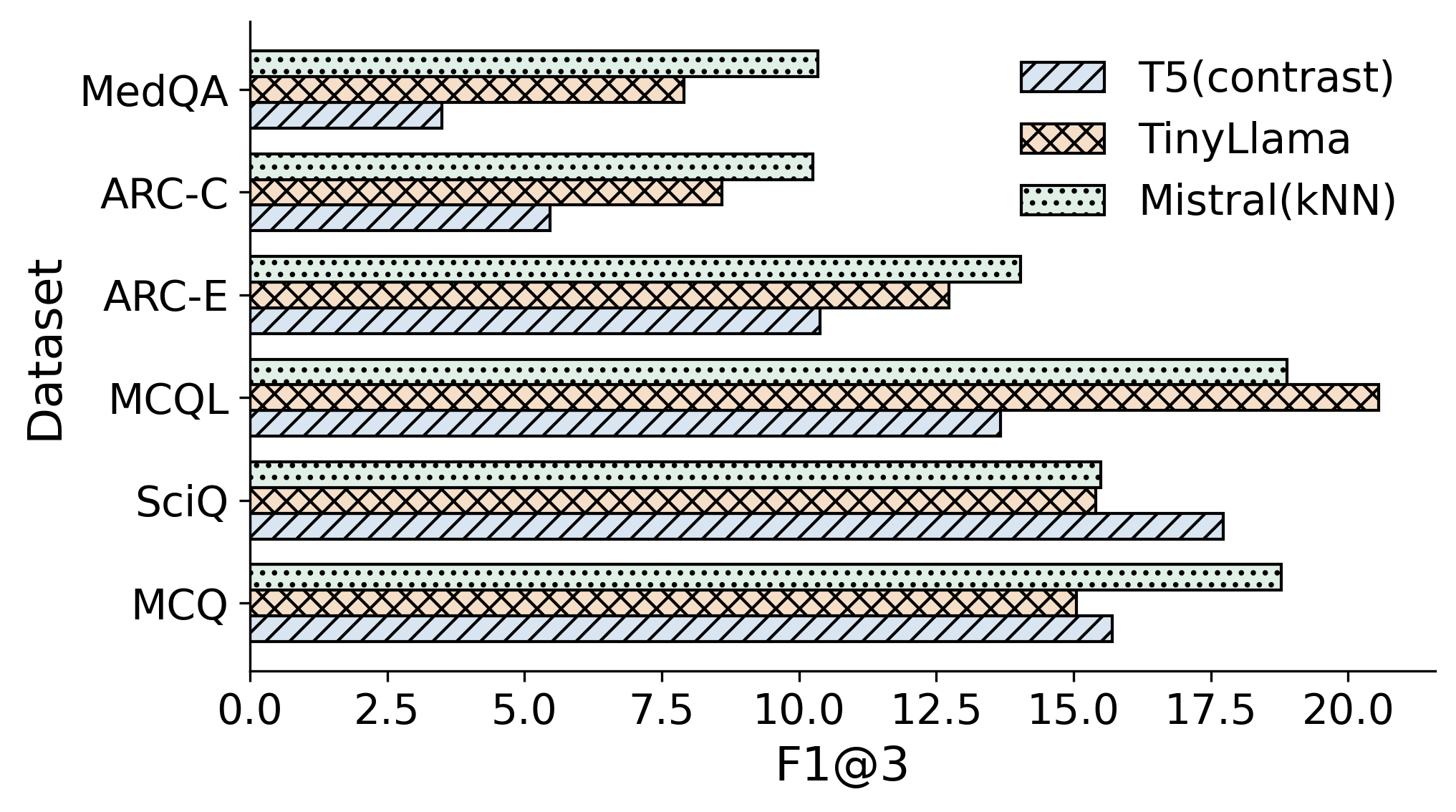}
 \vspace{-8mm}
     \caption{F1@3 of Mistral(k-NN) with recent models.}
    \label{fig:mistral}
    \vspace{-8mm}
\end{center}
\end{figure}

\begin{table*}[!t]
\centering
\resizebox{\textwidth}{!}{
\begin{tabular}{l|p{5cm}|p{7.2cm}|p{5cm}}
\hline\hline
\multicolumn{1}{c|}{\textbf{Model}}
&
\multicolumn{1}{c|}{\textbf{Example \hypertarget{example1}{(1)}}} &
\multicolumn{1}{c|}{\textbf{Example \hypertarget{example2}{(2)}}} &
\multicolumn{1}{c}{\textbf{Example \hypertarget{example3}{(3)}}} \\
\hline
\multirow{2}{*}{Question [\textcolor{blue}{Answer}] }

& Main source of carbohydrates for various animals is [\textcolor{blue}{starch}] 
& The common term for the chemical formula $H_2O$ is \textcolor{blue}{[water]}
& \textcolor{blue}{[humans]} has the greatest ability to learn. \\
\hline\hline

\textbf{Ground-truth} 
& [glycogen, agar, pectin] 
& [hydrogen, air, oxygen] 
& [apes, reptiles, rats] \\

\textbf{T5(base)} 
& [glucose, cellulose, \underline{starch}]
& [air, light, air]
& [animals, plants, animals] \\

\textbf{T5(multi-task)} 
& [glucose, cellulose, \underline{starch}]
& [air, ice, gas]
& [animals, animals, plants] \\

\textbf{T5(contrast)} 
& [proteins, fats, proteins] 
& [helium, nitrogen, carbon] 
& [animals, plants, animals] \\

\textbf{T5(beam)} 
& [\underline{starch}, cellulose, glucose] 
& [carbon, oxygen, gas] 
& [animals, mammals, dinosaurs] \\

\textbf{T5(clustering)} 
& [\underline{starch}, cellulose, sugar]
& [carbon, air, ions]
& [animals] \\

\textbf{GPT-3(clustering)} 
& [fat, fiber, maltose] 
& [H2O2, Hydroxide, Hydronium] 
& [elephants, monkeys, cats] \\

\textbf{TinyLlama} 
& [proteins, cellulose, sap] 
& [gas, liquid, heat] 
& [chickens, cars, fish] \\

\textbf{Mistral(k-NN)} 
& [cellulose, glycogen, glucose]
& [hydrogen peroxide, steam, ice] 
& [dogs, dolphins, parrots] \\

\textbf{GPT-3(random)} 
& [protein, cellulose, fat]
& [carbon dioxide, oxygen, hydrogen peroxide]
& [plants, animals, machines] \\

\textbf{GPT-3(k-NN)} 
& [cellulose, glycogen, fructose] 
& [hydrogen peroxide, carbon dioxide, ammonia] 
& [birds, dolphins, chimpanzees] \\

\textbf{GPT-3(COT-k-NN)} 
& [cellulose, glycogen, fructose] 
& [oxygen, hydrogen, carbon] 
& [dogs, birds, dolphins] \\

\hline\hline
\end{tabular}
}
\vspace{-2mm}
\caption{Examples of DG models. Underline indicates the generated distractor matches the correct answer.}
\label{tab:horizontal_examples}
\vspace{-5mm}
\end{table*}

\subsection{Analysis of Different Factors}
Table~\ref{tab:results_ablation} summarizes the analysis of several factors in ICL with unsupervised retrieval. We examine the impact of removing the answer from the few-shot examples (i.e., only the question and its distractors). 
The results show a substantial performance decline. The F1@3 score decreases from 19.69 to 15.57 on MCQ and drops more sharply from 15.08 to 3.61 on MedQA. This highlights the importance of full-context k-shot examples for ICL reasoning.

Then, we observed that MPNet sentence encoders may outperform MiniLM \cite{wang2020minilm} with k-NN retrieval for the DG. MPNet indicates better automatic metrics, particularly on challenging datasets such as ARC and MedQA. The F1@3 increases from 14.67 to 15.05 on ARC-challenge and 13.85 to 15.08 on MedQA.

We also report the performance with Mistral \cite{jiang2023mistral7b} large language model and k-NN retrieval. While Mistral achieves better automatic results compared to recent DG approaches, such as contrastive and instruction-tuned small models across most datasets, especially on harder reasoning benchmarks, as summarized in Figure~\ref{fig:mistral}, GPT-3 consistently outperforms it across all selected datasets, as detailed in Table \ref{tab:results_ablation}. This indicates the effectiveness of in-context learning for DG.

We evaluate the effect of varying the number of \(k\)-shot examples \(\{5, 7, 20, 30\}\) as detailed in Table \ref{tab:k-shot} in Appendix \ref{sec:k-shot_ablation}.  
We observe {\em new} SOTA automatic results across all selected datasets at \(k = 30\), outperforming recent approaches \cite{wang-etal-2023-distractor, yu-etal-2024-enhancing, alhazmi-etal-2025-fine}. 
This highlights the critical role of reasoning in DG task.

\subsection{Case Study} \label{sec:case_study}
Table \ref{tab:horizontal_examples} presents a case study of DG models, highlighting notable limitations of the CSG-DS framework. In Example \hyperlink{example1} {(1)}, T5 with beam search is prone to generate the correct answer as a distractor. In Example \hyperlink{example2} {(2)}, GPT-3 with clustering may select distractors (H2O2, Hydroxide, Hydronium) that lack the level of plausibility expected by experts.

The contrastive approach yields clear improvements over Text2Text baseline fine-tuning and multi-task learning, yet it still struggles to generate competitive distractors. As shown in Examples~\hyperlink{example1}{(1)} and~\hyperlink{example3}{(3)}, the contrastive approach exhibits repetitive outputs and limited reasoning. While Example~\hyperlink{example2}{(2)} contains more in-context distractors (helium, nitrogen, carbon) than out-of-context distractors (air, ice, gas), these options still lack the level of reasoning typically found in human-generated distractors and may provide limited challenge to examinees. Instruction-tuned models (e.g., TinyLlama) also exhibit notable issues, including generating out-of-context distractors (gas, liquid, heat), as shown in Example~\hyperlink{example2}{(2)}, or highly implausible options (chickens, cars, fish), as outlined in Example~\hyperlink{example3}{(3)}.

The ICL framework with k-NN retrieval generates more human-like distractors in all cases. In Example~\hyperlink{example1}{(1)}, the distractors (cellulose, glycogen, fructose) are conceptually aligned with the question-answer pair, introducing a meaningful knowledge challenge.  In Example~\hyperlink{example2}{(2)}, the generated distractors correspond to related chemical compounds (\(H_{2}O_{2}\), \(CO_{2}\), \(NH_{3}\)) rather than the target (\(H_{2}O\)), while COT augmentation further refines the output as human-like reasoning distractors (oxygen, hydrogen, carbon). Example~\hyperlink{example3}{(3)} shows that k-NN retrieval generates distractors that are more closely aligned with the ground truth than those generated by random retrieval.

\vspace{-1mm}
\section{Conclusion} 
\label{sec:conclusion}
\vspace{-1mm}
We present an in-context learning (ICL) framework that leverages the reasoning capabilities of large language models (LLMs) to generate distractors that more closely reflect human reasoning. We further incorporate chain-of-thought augmentation to generate both distractors and their associated rationales. This framework yields substantial improvements across challenging domains and varying distractor lengths. Both automatic and human evaluations show that ICL achieves state-of-the-art performance, generating more plausible, diverse, and contextually aligned distractors than recent fine-tuning and contrastive Text2Text methods. 

\section*{Limitations}
While in-context learning (ICL) with large language models (LLMs) demonstrates strong capability in generating contextually aligned and human-like distractors, it remains susceptible to hallucination. In particular, zero-shot prompting is more prone to undesired behaviors, such as repeating the question stem or leaking the correct answer among the distractors. Although few-shot prompting alleviates these issues to some extent, it does not fully eliminate them. Moreover, different LLMs exhibit distinct generation behaviors. For instance, models such as Mistral tend to generate long lists of distractors, whereas GPT-3 more consistently adheres to the intended output format.

Also, distractor evaluation still largely relies on automatic token-level matching against ground-truth distractors. Although ICL generates substantially fewer distractors that exactly match the ground-truth answer for the improvement of reliability, it remains challenging to validate the correctness of the generated distractors that may constitute plausible but valid answer options. This limitation highlights the difficulty of reliably distinguishing high-quality distractors from semantically acceptable answers using surface-level evaluation alone.

Despite these limitations, our findings consistently underscore the importance of explicit reasoning in the distractor generation task. Reasoning-aware in-context learning generates more plausible, diverse, and contextually grounded distractors than non-reasoning baselines. We hope that this work encourages the research community to further investigate robust output control, validation mechanisms, and reasoning-aware constraints for in-context learning–based distractor generation.

 \bibliography{custom}

\clearpage  
\appendix

\section{K-Nearest Neighbor Retrieval} \label{sec:k-NN_app}
The following algorithm 
describes the k-nearest neighbor (k-NN) retrieval
used to select in-context examples.
Algorithm~\ref{alg:k-NN_retrieval} presents the
in-context learning 
framework for distractor generation, where the model generates distractors
$\hat{y}$ and can optionally incorporate chain-of-thought (COT) 
augmentation to generate $\hat{z}$, consisting of both distractors $\hat{y}$ and its rationales $\hat{r}$ for a given input 
$x_{\text{test}}$.

\begin{algorithm}
\caption{In-Context Learning with k-NN-based Retrieval and Optional COT Augmentation}
\label{alg:k-NN_retrieval}
\textbf{Require:} Test input $x_{\text{test}}$; training set
$\mathcal{D}_T=\{(x_i, y_i [, r_i])\}_{i=1}^N$, where $r_i$ denotes an optional
chain-of-thought rationale available only when COT augmentation is enabled;
sentence encoder $\mu_\theta(\cdot)$;
number of few-shot examples $k$;
binary flag $\gamma \in \{0,1\}$, where $\gamma=0$ corresponds to
in-context learning and $\gamma=1$ activates COT augmentation. \\
1: $v_{\text{test}} \leftarrow \mu_\theta(x_{\text{test}})$ \\
2: \textbf{for each} $(x_i, y_i [, r_i]) \in \mathcal{D}_T$ \textbf{do} \\
3: \quad $v_i \leftarrow \mu_\theta(x_i)$ \\
4: \quad $s_i \leftarrow \frac{v_i^\top v_{\text{test}}}{\|v_i\|_2 \|v_{\text{test}}\|_2}$ \\
5: \textbf{end for} \\
6: Select indices $\mathcal{I}$ of the top-$k$ examples with highest similarity scores \\
7: Retrieve examples $\mathcal{C}=\{(x_i, y_i [, r_i]) \mid i \in \mathcal{I}\}$ \\
8: \textbf{if} $\gamma = 1$ \textbf{then} \\
9: \quad $C \leftarrow e_{r_1} \oplus \cdots \oplus e_{r_k}$ \\
10: \quad $\hat{z}=(\hat{y}, \hat{r}) \leftarrow \text{LLM}([C; x_{\text{test}}])$ \\
11: \textbf{else} \\
12: \quad $C \leftarrow e_{1} \oplus \cdots \oplus e_{k}$ \\
13: \quad $\hat{y} \leftarrow \text{LLM}([C; x_{\text{test}}])$ \\
14: \textbf{end if}
\end{algorithm}

\section {Datasets and Domains} \label{sec:datasets_domains}
We evaluate our approach on six multiple-choice question datasets across science, general knowledge, and medical domains.

\vspace{1mm}
The \noindent\textbf{Science} domain includes three main datasets.
{\em SciQ} \cite{welbl-etal-2017-crowdsourcing} is a crowd-sourced dataset with distractors and questions averaging 1.6 and 14.5 tokens, respectively. 
{\em MCQL} \cite{liang-etal-2018-distractor} is a Web-crawled college-level science dataset with distractors averaging 1.2 tokens and questions 9.4 tokens.
{\em ARC} \cite{clark2018think} covers elementary science questions at Easy and Challenge levels, with question lengths ranging from 21.4 to 24.7 tokens and distractors from 4.1 to 5.5 tokens. Examples~\hyperlink{exampleb1}{(B.1)} and 
Example~\hyperlink{exampleb2}
{(B.2)} illustrate two examples from SciQ and ARC-Challenge, respectively:

\begin{quote}
\begin{footnotesize}
\hypertarget{exampleb1}{(B.1)}
\textbf{Stem:} \textit{What is a disease caused by the same virus that causes chicken pox? }\\
\textbf{Distractors:} \textit{a) gout, b) diabetes, c) hepatitis}\\
\textbf{Answer:} \textit{shingles}
\end{footnotesize}
\end{quote}

\begin{quote}
\begin{footnotesize}
\hypertarget{exampleb2}{(B.2)}
\textbf{Stem:} \textit{Which is a fact about penguins? }\\
\textbf{Distractors:} \textit{a) Penguins are fierce competitors, b)  Penguins are some of the most beautiful birds, c) Penguins make great pets}\\
\textbf{Answer:} \textit{Penguins can live in climates with freezing temperatures}
\end{footnotesize}
\end{quote}

\vspace{1mm}
The \noindent\textbf{General Knowledge} domain refers to the {\em MCQ} dataset\footnote{\url{https://github.com/DRSY/DGen}} \cite{ren2021knowledge} that covers diverse domains including science, vocabulary, commonsense, and trivia. It contains fill-in-the-blank questions where we replace  “**blank**”  with a \texttt{[MASK]} token. MCQ features relatively short questions with 19.5 tokens on average and single-token distractors, as presented in Example~\hyperlink{exampleb3}{(B.3)}:

\begin{quote}
\begin{footnotesize}
\hypertarget{exampleb3}{(B.3)}
\textbf{Stem:} \textit{the only known planet with large amounts of water is [MASK] }\\
\textbf{Distractors:} \textit{a) saturn, b) jupiter, c) mars}\\
\textbf{Answer:} \textit{earth}
\end{footnotesize}
\end{quote}

\vspace{1mm}
The \noindent\textbf{Medical} domain refers to the {\em MedQA}  dataset \cite{jin2021disease} that is constructed from real medical licensing examinations. It contains substantially longer questions with 116.6 tokens on average and multi-token distractors with 3.5 tokens on average, as shown in Example~\hyperlink{exampleb4}{(B.4)}, making it challenging dataset for the DG task.

\begin{quote}
\begin{footnotesize}
\hypertarget{exampleb4}{(B.4)}
\textbf{Stem:} \textit{The parents of a 14-year-old patient are concerned and have questions about the use of insulin for their son's recently diagnosed type 1 diabetes. The patient has developed an upper respiratory infection while at school. He is coughing and has a runny nose. His temperature is $37.8\,^{\circ}\mathrm{C}\ (100.2\,^{\circ}\mathrm{F})$ and vital signs are within normal limits. Physical examination is unremarkable. Which of the following modifications to his insulin regimen would you recommend to this patient and his parents?}\\
\textbf{Distractors:} \textit{a) Reduce the insulin dose, b)  Continue same regimen, c) Hold insulin until the patient gets better, d) Increase the insulin dose to double}\\
\textbf{Answer:} \textit{Increase the frequency of blood glucose checks}
\end{footnotesize}
\end{quote}

\section{Prompt Template} \label{sec:templates}
We conduct in-context learning by conditioning the model on a given context \(C\), which consists of \(k\) example pairs \(\{(x_i, y_i)\}_{i=1}^{k}\) selected via a
retrieval mechanism and formatted using structured prompt templates. As illustrated in Example~\hyperlink{examplec1}{(C.1)}, the prompt begins with the \(k\) retrieved examples, followed by a \texttt{[stop]} token and the test input \(x_{\text{test}}\).

\begin{quote}
\begin{footnotesize}
\hypertarget{examplec1}{(C.1)}
{\bf Question}: \textit{If an object is attracted to a magnet, the object is most likely made of [MASK]} \\
{\bf Answer}: \textit{metal} \\
{\bf Distractor1}: \textit{wood} \\
{\bf Distractor2}: \textit{plastic} \\
{\bf Distractor3}: \textit{cardboard} \\
{\dots}\\
\texttt{[stop]}\\
{\bf Question}:  \textit{sugars are broken down into in your digestive system [MASK]}\\
{\bf Answer}: \textit{glucose}
\end{footnotesize}
\end{quote}

We also follow a structured prompt template for chain-of-thought (COT) augmentation, based on a given context \(C\) that includes \(k\) few-shot examples \(\{(x_i, y_i, r_i)\}_{i=1}^{k}\) selected via a retrieval mechanism. As presented in Example~\hyperlink{examplec2}{(C.2)}, the prompt template starts with \(k\) examples, followed by a \texttt{[stop]} token and then the test input \(x_{\text{test}}\).

\begin{quote}
\begin{footnotesize}
\hypertarget{examplec2}{(C.2)}
{\bf Question}: \textit{If an object is attracted to a magnet, the object is most likely made of [MASK]}\\
{\bf Answer}: \textit{metal} \\
{\bf Distractor1}: \textit{wood} \\
{\bf Distractor2}: \textit{plastic} \\
{\bf Distractor3}: \textit{cardboard} \\
{\bf Rationale} : \textit{The fact that the object is attracted to a magnet indicates that it contains magnetic properties. The most common materials that exhibit magnetic properties are metals, such as iron, nickel, and cobalt. Therefore, the object is most likely made of metal if it is attracted to a magnet. Wood, plastic, and cardboard do not typically exhibit magnetic properties, so they are not likely to be the correct answer.}\\
{\dots}\\
\texttt{[stop]}\\
{\bf Question}:  \textit{sugars are broken down into in your digestive system [MASK]}\\
{\bf Answer}: \textit{glucose}
\end{footnotesize}
\end{quote}

We use a zero-shot prompt template, as illustrated in Example~\hyperlink{examplec3}{(C.3)}, which maps a given question-answer input \(x\) and an explicit instruction to generate distractor outputs \(y\).

\begin{quote}
\begin{footnotesize}
\hypertarget{examplec3}{(C.3)}
{\bf Generate three incorrect distractors for the following question.}\\
{\bf Question}: \textit{sugars are broken down into in your digestive system [MASK]}\\
{\bf Answer}: \textit{glucose}
\end{footnotesize}
\end{quote}

For candidate generation within the candidate generation and selection framework, we use the prompt template outlined in Example~\hyperlink{examplec4}{(C.4)} to generate ten candidate distractors. Few-shot examples are selected using k-NN-based retrieval.

\begin{quote}
\begin{footnotesize}
\hypertarget{examplec4}{(C.4)}
{\bf Question}: \textit{If an object is attracted to a magnet, the object is most likely made of [MASK]} \\
{\bf Answer}: \textit{metal} \\
{\bf Distractor1}: \textit{wood} \\
{\bf Distractor2}: \textit{plastic} \\
{\bf Distractor3}: \textit{cardboard} \\
{\dots}\\
\texttt{[stop]}\\
{\bf Generate ten incorrect distractors for the following question and answer}\\
{\bf [Template] \\
           Distractor1: \\
           Distractor2: \\
           Distractor3:\\}
{\bf Question}:  \textit{sugars are broken down into in your digestive system [MASK]}\\
{\bf Answer}: \textit{glucose}
\end{footnotesize}
\end{quote}

\section {Models and Implementation Details} \label{sec:models}
We implement all models using the Hugging Face Transformers framework \cite{wolf-etal-2020-transformers} and conduct experiments on two NVIDIA Tesla P100 GPUs. We also detail the implementation settings and training configurations for each model utilized in distractor generation and question answering.

\subsection{Distractor Generation Models}
We conduct distractor generation experiments using T5, TinyLlama, GPT-3, and Mistral models. The following describes the experimental settings for each model.

\vspace{1mm}
\noindent\textbf{T5}: We optimize 
the \emph{T5-base} model by using AdamW with an initial learning rate of 1e-4 for 10 epochs and a batch size of 4 within a candidate generation framework as well as the Text2Text architecture.  We integrate a multi-task learning setup with two objectives: question answering and distractor generation. For the contrastive-based variant, we adopt a two-stage training. The model is first fine-tuned using a standard generation objective, followed by additional fine-tuning with a combined loss consisting of the InfoNCE contrastive loss and the generation loss. The temperature $\tau$ is fixed at 0.1, and mean pooling is applied to obtain fixed-size embedding representations. For clustering, we apply agglomerative clustering\footnote{\url{https://scikit-learn.org/stable/modules/generated/sklearn.cluster.AgglomerativeClustering.html}} using Euclidean distance, with a distance threshold of 1.2. The representative element from each resulting cluster is then selected as the final distractor.

One important factor is the training cost of pre-trained language models such as T5 model. Within the Text2Text architecture, Example~\hyperlink{exampled1}{(D.1)} reports the training time required to fine-tune the two-stage contrastive Text2Text model across datasets, ranging from under 30 minutes on ARC-Challenge to over 5 hours on MedQA. Example~\hyperlink{exampled2}{(D.2)} presents the training time associated with multi-task learning, which consistently incurs slightly higher computational cost across all datasets compared to the contrastive-based approach. These results indicate that, while the contrastive method requires marginally lower computational cost than multi-task learning, it also achieves better performance on the distractor generation task, especially in the Text2Text architecture.

\begin{quote}
\begin{footnotesize}
\hypertarget{exampled1}{(D.1)}
{\bf Training Time:}  \textit{Contrastive Text2Text }\\
{\bf MCQ}:  \textit{45 min 4 s} \\
{\bf SciQ}:  \textit{4 h 28 min 52 s} \\
{\bf MCQL}:   \textit{2 h 23 min 23 s}\\
{\bf ARC-Easy}:  \textit{55 min 19 s}\\
{\bf ARC-Challenge}:  \textit{28 min 30 s}\\
{\bf MedQA}:  \textit{5 h 12 s}\\
\end{footnotesize}
\end{quote}
 
\begin{quote}
\begin{footnotesize}
\hypertarget{exampled2}{(D.2)}
{\bf Training Time:} \textit{Multi-task Learning} \\
{\bf MCQ}: \textit{  57 min 27 s    }\\
{\bf SciQ}: \textit{ 5 h 51 min 11 s   }\\
{\bf MCQL}: \textit{ 2 h 55 min 54 s    }\\
{\bf ARC-Easy}: \textit{ 1 h 11 min 49 s     }\\
{\bf ARC-Challenge}: \textit{ 37 min 22 s    }\\
{\bf MedQA}: \textit{ 5 h 17 min 22 s    }\\
\end{footnotesize}
\end{quote}

Additionally, the results in Example~\hyperlink{exampled3}{(D.3)} indicate that fine-tuning T5 for candidate generation is computationally more expensive, requiring substantially longer training times across the selected datasets than Text2Text-based methods. For instance, training on SciQ and MedQA requires over 8 hours and 10 hours, respectively, while even mid-sized datasets such as MCQL and ARC-Easy demand more than 4 hours and nearly 2 hours of training. In contrast, Text2Text approaches complete training in significantly less time, highlighting the higher computational overhead associated with candidate generation using T5.

\begin{quote}
\begin{footnotesize}
\hypertarget{exampled3}{(D.3)}
{\bf Training Time:}  \textit{T5 Candidate Generation }\\
{\bf MCQ}: \textit{ 1 h 31 min 17 s  }\\
{\bf SciQ}: \textit{ 8 h 17 min 35 s }\\
{\bf MCQL}: \textit{ 4 h 25 min }\\
{\bf ARC-Easy}: \textit{ 1 h 44 min 56 s }\\
{\bf ARC-Challenge}: \textit{ 54 min 7 s }\\
{\bf MedQA}: \textit{ 10 h 8 min 52 s }\\
\end{footnotesize}
\end{quote}

\noindent\textbf{TinyLlama}: 
We fine-tune \emph{TinyLlama-1.1B-Chat-v1.0}, using instruction tuning with Low-Rank Adaptation (LoRA) \cite{hu2022lora}. 
The model is trained with prompts structured as question-answer pairs, where the model generates three distractors per question. We apply LoRA with rank \(r=8\), scaling factor \(\alpha=16\), and a dropout rate of 0.05, targeting the projection layers \texttt{q\_proj}, \texttt{k\_proj}, \texttt{v\_proj}, \texttt{o\_proj}, \texttt{gate\_proj}, \texttt{up\_proj}, and \texttt{down\_proj}. The model is optimized using AdamW with a learning rate of 2e-4 for 10 epochs, a batch size of 4, and mixed-precision (FP16) training. For inference, we apply greedy decoding with a maximum generation length of 50 tokens.

Fine-tuning large language models through instruction tuning requires substantially more computational time than all recent approaches, including pre-trained models such as T5. As shown in Example~\hyperlink{exampled4}{(D.4)}, instruction tuning with TinyLlama language model can exceed 24 hours on SciQ and more than two days on MedQA, highlighting its significant computational overhead.

\begin{quote}
\begin{footnotesize}
\hypertarget{exampled4}{(D.4)}
{\bf Training Time:}  \textit{Instruction Tuning}\\
{\bf MCQ}: \textit{ 4 h 43 min 44 s }\\
{\bf SciQ}: \textit{ 1 day 23 h 52 min 18 s }\\
{\bf MCQL}: \textit{ 18 h 46 min 47 s }\\
{\bf ARC-Easy}: \textit{ 6 h 26 min 59 s }\\
{\bf ARC-Challenge}: \textit{ 2 h 20 min 38 s }\\
{\bf MedQA}: \textit{2 days 1 h 32 min 43 s }\\
\end{footnotesize}
\end{quote}

\begin{table*}[t]
\centering
\resizebox{\textwidth}{!}{%
\begin{tabular}{c|cccccc|cccccc}
\hline\hline
 & \multicolumn{6}{c|}{\textbf{T5 Accuracy (\%)}} 
 & \multicolumn{6}{c}{\textbf{BART Accuracy (\%)}} \\

\textbf{Dataset} 
 & MCQ & SciQ & MCQL & ARC-E & ARC-C & MedQA
 & MCQ & SciQ & MCQL & ARC-E & ARC-C & MedQA \\
\hline
Ground-Truth     
 & 48.26 & 68.00 & 34.00 & 40.45 & 27.13 & 21.76
 & 30.12 & 52.50 & 27.33 & 29.38 & 19.80 & 20.42 \\

T5 (beam)        
 & 56.76 & 64.70 & 49.17 & 38.55 & 36.52 & 32.68
 & 40.12 & 51.70 & 43.00 & 39.35 & 31.91 & 31.81 \\

GPT-3 (clustering)
 & 42.08 & 59.60 & 39.00 & 40.74 & 31.66 & 29.69
 & 26.25 & 47.70 & 34.83 & 31.40 & 28.50 & 28.04 \\

T5 (contrast)    
 & 47.10 & 62.30 & 42.67 & 38.72 & 32.51 & 26.71
 & 35.14 & 52.70 & 36.67 & 35.69 & 25.85 & 24.19 \\

GPT-3 (k-NN)      
 & 41.31 & 50.30 & 30.83 & 36.53 & 25.43 & 21.05
 & 25.87 & 39.80 & 26.50 & 26.22 & 20.31 & 20.03 \\

GPT-3 (COT-k-NN)  
 & 36.29 & 50.20 & 30.83 & 35.56 & 27.30 & 19.95
 & 22.78 & 40.90 & 26.33 & 27.99 & 20.90 & 19.25 \\

TinyLlama        
 & 50.19 & 63.80 & 37.83 & 41.20 & 31.14 & 24.12
 & 26.64 & 52.60 & 33.00 & 29.50 & 21.59 & 25.22 \\
\hline\hline
\end{tabular}%
}
\caption{Accuracy comparison (\%) of distractor generation approaches using the T5 and BART across datasets.}
\label{tab:t5_bart_accuracy}
\end{table*}

\noindent\textbf{GPT-3}: We use the \texttt{gpt-3.5-turbo} model within an in-context learning framework, with \(k=5\) shots, temperature set to 0, a maximum generation length of 350 tokens, and top-\(p=1\), corresponding to greedy decoding. For k-NN-based retrieval, we employ a pre-trained MPNet encoder \cite{reimers-gurevych-2019-sentence} and compute cosine similarity between vectorized textual representations of test and training instances. The implementation leverages the OpenAI API with Azure OpenAI support to generate model responses. Prompts are processed in configurable batches of 20 per request.

Although GPT-3 is not an open-source large language model, it exhibits strong distractor generation performance while achieving significantly lower inference time, as shown in Example~\hyperlink{exampled5}{(D.5)}. For instance, inference completes within 2–10 minutes across all datasets, including less than 3 minutes on MCQ and ARC-Challenge, demonstrating its efficiency compared to fine-tuning-based approaches, including candidate generation and selection frameworks, Text2Text models, and instruction-tuned models.

\begin{quote}
\begin{footnotesize}
\hypertarget{exampled5}{(D.5)}
{\bf Inference Time:}  \textit{GPT-3 ICL}\\
{\bf MCQ}: \textit{  2 min 6 s}\\
{\bf SciQ}: \textit{ 9 min 29 s}\\
{\bf MCQL}: \textit{ 4 min 38 s}\\
{\bf ARC-Easy}: \textit{ 5 min 32 s}\\
{\bf ARC-Challenge}: \textit{ 2 min 46 s}\\
{\bf MedQA}: \textit{ 10 min 9 s}\\
\end{footnotesize}
\end{quote}

\vspace{1mm}
\noindent\textbf{Mistral}: We explore the \emph{Mistral-7B-Instruct} model via Ollama within the in-context learning framework, with \(k=5\) shots, temperature set to 0, top-\(p=1\), and \texttt{num\_predict} = 1500, corresponding to greedy decoding. The model is deployed locally through the Ollama interface. Unlike GPT-3, we select only the first three generated distractors from the output results, as the Mistral large language model tends to generate long distractor lists.

\begin{quote}
\begin{footnotesize}
\hypertarget{exampled6}{(D.6)}
{\bf Inference Time:}  \textit{Mistral ICL}\\
{\bf MCQ}: \textit{ 37 min 59 s }\\
{\bf SciQ}: \textit{ 1 h 24 min 51 s  }\\
{\bf MCQL}: \textit{ 30 min 41 s  }\\
{\bf ARC-Easy}: \textit{ 1 h 55 min 44 s   }\\
{\bf ARC-Challenge}: \textit{48 min 16 s}\\
{\bf MedQA}: \textit{  1 h 28 min 48 }\\
\end{footnotesize}
\end{quote}

Compared to the fine-tuning models, Mistral under the in-context learning paradigm eliminates the need for costly training, enabling immediate and flexible deployment. However, this advantage comes at the cost of higher inference latency, as shown in Example~\hyperlink{exampled6}{(D.6)}. Although fine-tuned approaches incur substantial training costs (Examples~\hyperlink{exampled1}{(D.1)}–\hyperlink{exampled4}{(D.4)}), they benefit from significantly faster inference once training is completed. This highlights a clear trade-off between training efficiency and inference efficiency.

 \begin{figure*}[!tb]
\begin{center}
    \includegraphics[width=1\textwidth]{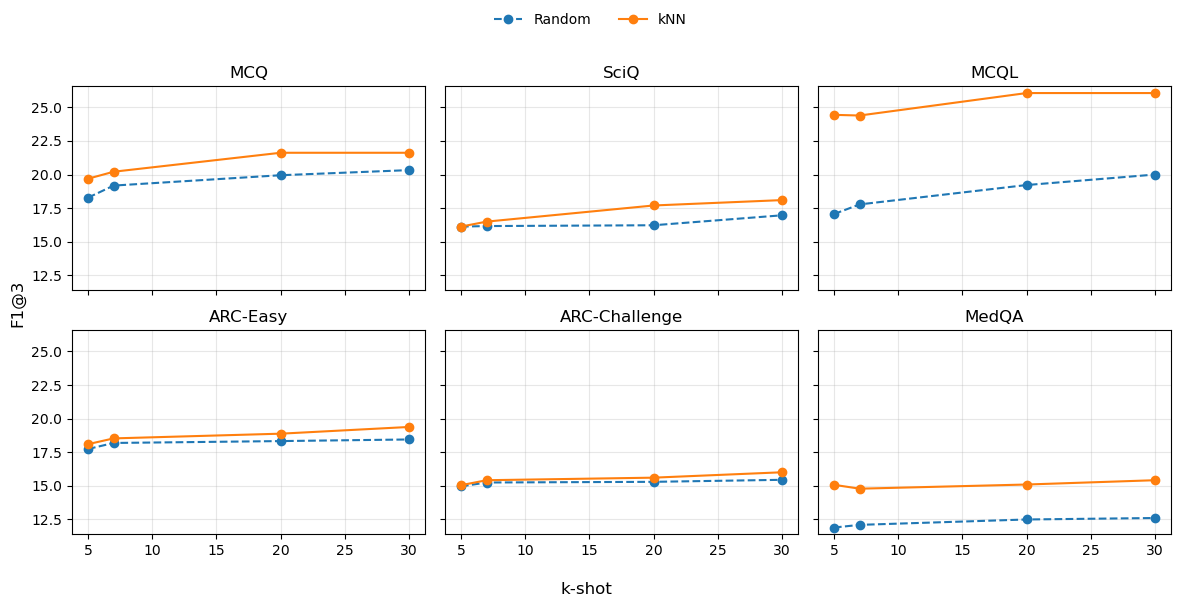}
        \vspace{-5mm}
    \caption{{F1@3 score comparison with varying number of few-shot ($k \in \{5,7,20,30\}$) across datasets.}
    } 
    \label{fig:k_shot_fig}
    \vspace{-2mm}
\end{center}
\end{figure*}

\subsection{Question Answering Models}
We employ two encoder–decoder models, T5 \cite{raffel2020exploring} and BART \cite{lewis-etal-2020-bart}, as question-answering models. The models {\em t5-base} and {\em facebook/bart-base} are fine-tuned in a supervised manner, conditioned on the given question and its associated options, which randomly include the generated distractors and the correct answer. The input is prefixed with the labels "\texttt{question:}" and "\texttt{options:}" before the corresponding content.

The models are optimized using AdamW with a learning rate of 1e-4, weight decay of 0.01, trained for 10 epochs with a batch size of 4. We set maximum input length to 512 tokens and maximum target length to 128 tokens. During inference, we use greedy decoding with early stopping and a maximum generation length of 50 tokens. The model is evaluated based on exact match accuracy between generated and ground-truth answers.

Let $\mathcal{D} = \{(a_i, \hat{a}_i)\}_{i=1}^{N}$ denote a set of $N$ evaluation instances,
where $a_i$ is the ground-truth answer and $\hat{a}_i$ is the generated answer.
We define the indicator function $\mathbb{I}(\cdot)$ as:
\begin{equation}
\mathbb{I}(a_i = \hat{a}_i) =
\begin{cases}
1, & \text{if } a_i = \hat{a}_i, \\
0, & \text{otherwise}.
\end{cases}
\end{equation}
the accuracy is computed as:
\begin{equation}
\frac{1}{N} \sum_{i=1}^{N} \mathbb{I}(a_i = \hat{a}_i) \times 100
\end{equation}
where both $a_i$ and $\hat{a}_i$ are normalized by lowercasing and trimming whitespace. 
Table~\ref{tab:t5_bart_accuracy} reports accuracy results for distractor generation approaches.

\section {k-Shot Ablation} \label{sec:k-shot_ablation}

Table~\ref{tab:k-shot} reports the automatic evaluation metrics for analyzing the effect of varying the number of \emph{k}-shot examples \(\{5, 7, 20, 30\}\) across all selected datasets. The performance 
shows a clear and consistent trend. The automatic results steadily improve as \(k\) increases for both random and k-NN retrieval. This indicates that richer in-context examples substantially benefit distractor generation in aligning closely with human-labeled distractors.

Figure~\ref{fig:k_shot_fig} summarizes the impact of varying the number of \emph{k}-shot examples on the F1@3 score. k-NN-based retrieval consistently outperforms random retrieval across all datasets, with particularly significant improvements observed on MCQL and MedQA. Specifically, the F1@3 score increases from 20.00 to 26.06 on MCQL and from 12.61 to 15.42 on MedQA when using k-NN selected examples with \(k = 30\). This highlights the importance of selecting semantically relevant examples rather than relying on randomly sampled contexts.

Furthermore, the highest F1@3 scores are consistently achieved when \(k = 30\) across all selected datasets, indicating \emph{new} state-of-the-art (SOTA) automatic results compared to recent approaches, as reported in Table~\ref{tab:results_main}. Under k-NN-based retrieval, increasing \(k\) from 5 to 30 improves F1@3 from 19.69 to 21.62 on MCQ, from 16.53 to 18.10 on SciQ, from 24.44 to 26.06 on MCQL, from 18.10 to 19.38 on ARC-Easy, from 15.05 to 16.01 on ARC-Challenge, and from 15.08 to 15.42 on MedQA. These results further demonstrate that leveraging k-NN-based retrieval with larger \emph{k}-shot contexts provides more informative in-context examples, leading to consistent improvements in distractor generation performance across varying distractor token lengths and diverse domains.

\begin{table*}[tb!]
\centering
\resizebox{0.85\textwidth}{!}{
\setlength{\tabcolsep}{6pt}
\begin{tabular}{c|c|l|c|c|c|c|c}
\hline\hline
\textbf{Dataset} & \textbf{$k$} & \textbf{Model}
& \textbf{P@1} & \textbf{R@1} & \textbf{F1@3} & \textbf{MRR}
& \textbf{NDCG@3} \\
\hline

\multirow{10}{*}{\textbf{MCQ}}
& 5  & \multirow{5}{*}{GPT-3 (few-shot-random)} & 25.87 & 8.62 & 18.28 & 33.72 & 40.68 \\
& 7  &                                          & 27.41 & 9.14 & 19.18 & 35.20 & 42.08 \\
& 20 &                                          & 30.50 & 10.17& 19.95 & 37.77 & 44.11\\
& 30 &                                          & 30.89 & 10.30& 20.33 & 38.48 & 45.27 \\
\cline{2-8}
& 5  & \multirow{5}{*}{GPT-3 (few-shot-k-NN)}    & 29.73 & 9.91  & 19.69 & 36.62 & 42.37 \\
& 7  &                                          & 30.12 & 10.04 & 20.21 & 36.49 & 41.44 \\
& 20 &                                          & {\bf 32.05} & {\bf 10.68} & {\bf 21.62} & {\bf 39.32} &  45.26 \\
& 30 &                                          & 30.12 & 10.04 & 21.62 & 38.42 & {\bf 45.29}  \\
\hline

\multirow{10}{*}{\textbf{SciQ}}
& 5  & \multirow{5}{*}{GPT-3 (few-shot-random)} & 25.10 & 8.37 & 16.13 & 30.17 & 34.57 \\
& 7  &                                          & 24.90 & 8.30 & 16.17 & 29.92 & 34.24  \\
& 20 &                                          & 25.00 & 8.33 & 16.23 & 30.12 & 34.36  \\
& 30 &                                          & 25.60 & 8.53 & 16.97 & 31.07 & 35.65 \\
\cline{2-8}
& 5  & \multirow{5}{*}{GPT-3 (few-shot-k-NN)}   & 24.90 & 8.30 & 16.53 & 30.45 & 35.24 \\
& 7  &                                         & 24.70 & 8.23 & 16.50 & 30.37 & 35.32 \\
& 20 &                                         & 25.20 & 8.40 & 17.70 & 31.63 & 37.25 \\
& 30 &                                         & {\bf 25.90} & {\bf 8.63} & {\bf 18.10} & {\bf 32.18} & {\bf 37.61} \\
\hline

\multirow{10}{*}{\textbf{MCQL}}
& 5  & \multirow{5}{*}{GPT-3 (few-shot-random)}  & 30.17 & 10.06 & 19.06 & 34.36 & 37.98  \\
& 7  &                                           & 29.00 & 9.67  & 17.78  & 32.53  &  35.48 \\
& 20 &                                           & 31.83 & 10.61 & 19.22 & 35.22 & 37.98 \\
& 30 &                                           & 33.83 & 11.28 & 20.00 & 37.17 & 39.88 \\
\cline{2-8}
& 5  & \multirow{5}{*}{GPT-3 (few-shot-k-NN)}   & 35.33 & 11.78 & 24.44 & 39.61 & 42.81 \\
& 7  &                                         & 36.00 & 12.00 & 24.39  & 40.36 & 43.86 \\
& 20 &                                         & 37.50 & 12.50 & {\bf 26.06} & 42.08  & {\bf 45.69} \\
& 30 &                                         & {\bf 38.17} & {\bf 12.72} & 26.06 & {\bf 42.17} & 45.19  \\
\hline

\multirow{10}{*}{\textbf{ARC-Easy}}
& 5  & \multirow{5}{*}{GPT-3 (few-shot-random)}  & 23.19 & 7.75  & 17.74 & 28.37 & 32.41  \\
& 7  &                                           & 23.95 & 8.00 & 18.19  & 29.03 & 32.95  \\
& 20 &                                           & 24.37 & 8.14 & 18.33 & 29.48 &  33.39 \\
& 30 &                                           & 23.86 & 7.98 & 18.25 & 29.15 & 33.29  \\
\cline{2-8}
& 5  & \multirow{5}{*}{GPT-3 (few-shot-k-NN)}   & 24.24 & 8.10 & 18.10 & 29.08 & 32.82  \\
& 7  &                                         & 24.66 & 8.24 & 18.53 & 29.63 & 33.39  \\
& 20 &                                         & 24.87 & 8.31 & 18.88 & 29.85 & 33.59 \\
& 30 &                                         & {\bf 25.93} & {\bf 8.66} & {\bf 19.38} & {\bf 30.77} & {\bf 34.37} \\
\hline

\multirow{10}{*}{\textbf{ARC-Challenge}}
& 5  & \multirow{5}{*}{GPT-3 (few-shot-random)}  & 21.08 & 7.07 & 14.96 & 24.82 & 27.75 \\
& 7  &                                           & 20.48 & 6.87 & 15.25 & 24.96 & 28.60  \\
& 20 &                                           & 20.90 & 7.01 & 15.30 & 25.24 & 28.76  \\
& 30 &                                           & 21.16 & 7.10 & 15.45 & 25.20 & 28.36 \\
\cline{2-8}
& 5  & \multirow{5}{*}{GPT-3 (few-shot-k-NN)}   & 20.48 & 6.87 & 15.05 & 24.06 & 26.80  \\
& 7  &                                         & 20.65 & 6.93 & 15.42 & 24.63 & 27.77  \\
& 20 &                                         & 20.48 & 6.87 & 15.61 & 24.64 & 27.88  \\
& 30 &                                         & {\bf 21.25} & {\bf 7.13} & {\bf 16.01} & {\bf 25.46} & {\bf 28.77} \\
\hline

\multirow{10}{*}{\textbf{MedQA}}
& 5  & \multirow{5}{*}{GPT-3 (few-shot-random)}  & 17.44 & 4.36 & 11.90 & 23.10 & 26.50  \\
& 7  &                                           & 18.22 & 4.56 & 12.10 & 23.46  & 26.44  \\
& 20 &                                           & 18.70 & 4.67 & 12.50& 24.09 &  27.18 \\
& 30 &                                           & 19.80 & 4.95 & 12.61 & 24.89 & 27.79  \\
\cline{2-8}
& 5  & \multirow{5}{*}{GPT-3 (few-shot-k-NN)}   & 20.90 & 5.22 & 15.08 & 27.44 & 30.52  \\
& 7  &                                         & 20.42 & 5.11 & 14.79 & 26.90 & 30.02  \\
& 20 &                                         & 21.45 &  5.36 & 15.10& 27.89 & 30.70 \\
& 30 &                                         & {\bf 22.00} & {\bf 5.50} & {\bf 15.42} & {\bf 28.45} & {\bf 31.52}  \\
\hline\hline

\end{tabular}
}
\vspace{-1mm}
\caption{Ablation results with varying number of few-shot examples
($k \in \{5,7,20,30\}$) across datasets.}
\label{tab:k-shot}
\vspace{-1mm}
\end{table*}

\section{Supplementary Examples}
To further support the case study presented in Table~\ref{tab:horizontal_examples}, we provide additional extended case studies in Tables~\ref{tab:T8}–\ref{tab:T12}.
These supplementary examples span a broader range of question types, domains, and distractor difficulty levels across the selected datasets.
We also report the generated distractors together with their corresponding rationales generated via the chain-of-thought augmentation.

Consistent with the main case study in Section \ref{sec:case_study}, the extended examples reveal the limitations of the candidate generation and selection framework, contrastive Text2Text, and instruction-tuned models, including answer leakage, repetitive distractors, and the generation of implausible distractors.

In contrast, the in-context learning framework, particularly when combined with k-NN based retrieval and COT augmentation, consistently generates more contextually and human-like distractors compared to recent approaches.
These extended examples indicate the robustness of in-context learning and highlight the importance of semantically relevant example selection and reasoning-aware prompting for effective distractor generation.

\begin{table*}[!t]
\centering
\small
\resizebox{\textwidth}{!}{
\begin{tabular}{p{0.18\textwidth} p{0.40\textwidth} p{0.44\textwidth}}
\hline\hline

\multicolumn{3}{p{0.95\textwidth}}{
\textbf{Example}: F.1 \par
\textbf{Dataset}: MCQL \par
\textbf{Question [\textcolor{blue}{Answer}]}:
Main source of carbohydrates for various animals is {[}\textcolor{blue}{starch}{]}
} \\
\hline\hline
\textbf{Model} & \textbf{Generated Distractors} & \textbf{Rationale} \\
\hline
Ground Truth                & [glycogen, agar, pectin]   & -- \\
T5(base)                    & [glucose, cellulose, \underline{starch}]   & -- \\
T5(multi-task)              & [glucose, cellulose, \underline{starch}]   & -- \\
T5(contrast)             & [proteins, fats, proteins]   & -- \\
T5(beam)                    & [\underline{starch}, cellulose, glucose]   & -- \\
T5(clustering)              & [\underline{starch}, cellulose, sugar]   & -- \\
GPT-3(clustering)           & [fat, fiber, maltose]     & -- \\
TinyLlama                   & [proteins, cellulose, sap]   & -- \\
Mistral(k-NN)                & [cellulose, glycogen, glucose] & -- \\
GPT-3(random)               & [protein, cellulose, fat]   & -- \\
GPT-3(k-NN)                  & [cellulose, glycogen, fructose]   & -- \\
GPT-3(COT-random)           & [cellulose, glucose, glycogen]
&[Starch is a complex carbohydrate found in plants and is a common source of carbohydrates for various animals. Cellulose is also a complex carbohydrate found in plants, but it is not easily digestible by many animals and is not a main source of carbohydrates for them. Glucose is a simple sugar and can be a source of carbohydrates, but it is not the main source for various animals. Glycogen is a form of stored glucose in animals, but it is not the main source of carbohydrates for them. Therefore, the correct answer is starch, as it is a common and easily digestible source of carbohydrates for various animals.] \\
GPT-3(COT-k-NN)              & [cellulose, glycogen, fructose] 
&  [The main source of carbohydrates for various animals is starch, which is a polysaccharide found in plants and serves as a source of energy. Cellulose is also a polysaccharide found in plants, but it is not easily digestible by many animals and is not their main source of carbohydrates. Glycogen is a form of carbohydrate storage in animals, not a source of carbohydrates from external sources. Fructose is a simple sugar found in fruits, but it is not the main source of carbohydrates for animals. Therefore, the correct answer is starch.]\\
\hline\hline

\multicolumn{3}{p{0.95\textwidth}}{
\textbf{Example}: F.2 \par
\textbf{Dataset}: MCQ \par
\textbf{Question [\textcolor{blue}{Answer}]}:
The common term for the chemical formula $H_2O$ is \textcolor{blue}{[water]}
} \\
\hline\hline
\textbf{Model} & \textbf{Generated Distractors} & \textbf{Rationale} \\
\hline
Ground Truth                & [hydrogen, air, oxygen]    & --\\
T5(base)                    & [air, light, air]   & -- \\
T5(multi-task)              & [air, ice, gas]   & -- \\
T5(contrast)             & [helium, nitrogen, carbon]    & -- \\
T5(beam)                    & [carbon, oxygen, gas]    & -- \\
T5(clustering)              & [carbon, air, ions]  & -- \\
GPT-3(clustering)           & [H2O2, Hydroxide, Hydronium]    & -- \\
TinyLlama                   & [gas, liquid, heat]    & -- \\
Mistral(k-NN)                & [hydrogen peroxide, steam, ice]  & --\\
GPT-3(random)               & [carbon dioxide, oxygen, hydrogen peroxide]  &  -- \\
GPT-3(k-NN)                  & [hydrogen peroxide, carbon dioxide, ammonia]    & -- \\
GPT-3(COT-random)           & [hydrogen peroxide, carbon dioxide, sodium chloride] 
&  [The chemical formula H2O represents water. The distractors hydrogen peroxide, carbon dioxide, and sodium chloride are incorrect as they represent different chemical compounds and do not match the given chemical formula H2O. Therefore, the correct answer is water as it matches the chemical formula H2O.]\\

GPT-3(COT-k-NN)              & [oxygen, hydrogen, carbon]
&  [The chemical formula H2O represents a specific compound. The common term for the chemical formula H2O is water. Oxygen, hydrogen, and carbon are not the common terms used to refer to the chemical formula H2O. Therefore, the correct answer is water.] \\
\hline\hline

\end{tabular}
}
\caption{Representative examples ({\bf F.1}, {\bf F.2}) of distractor generation models across selected datasets. Underline indicates cases where the generated distractor matches the correct answer.}
\label{tab:T8}
\end{table*}

\begin{table*}[!t]
\centering
\small
\resizebox{\textwidth}{!}{
\begin{tabular}{p{0.18\textwidth} p{0.40\textwidth} p{0.44\textwidth}}
\hline\hline

\multicolumn{3}{p{0.95\textwidth}}{
\textbf{Example}: F.3 \par
\textbf{Dataset}: MCQ \par
\textbf{Question [\textcolor{blue}{Answer}]}:
\textcolor{blue}{[humans]} has the greatest ability to learn. } \\
\hline\hline
\textbf{Model} & \textbf{Generated Distractors} & \textbf{Rationale} \\
\hline
Ground Truth                & [apes, reptiles, rats]    & -- \\
T5(base)                    & [animals, plants, animals]    & -- \\
T5(multi-task)              & [animals, animals, plants]   & -- \\
T5(contrast)             & [animals, plants, animals]   & -- \\
T5(beam)                    & [animals, mammals, dinosaurs]   & -- \\
T5(clustering)              & [animals]   & -- \\
GPT-3(clustering)           & [elephants, monkeys, cats]   & -- \\
TinyLlama                   & [chickens, cars, fish]   & -- \\
Mistral(k-NN)                & [dogs, dolphins, parrots]  & --\\
GPT-3(random)               & [plants, animals, machines]    & -- \\
GPT-3(k-NN)                  & [birds, dolphins, chimpanzees]   & -- \\
GPT-3(COT-random)           & [plants, animals, machines]
& [The question is asking about which entity has the greatest ability to learn. The answer is humans because humans are known for their advanced cognitive abilities and capacity for learning and adaptation. The distractors plants, animals, and machines are not as capable of learning as humans are, making them incorrect answers.] \\
GPT-3(COT-k-NN)              & [dogs, birds, dolphins]  
&  [The question is asking about the greatest ability to learn, which implies a high level of cognitive ability. Humans are known for their advanced cognitive abilities and capacity for learning, making them the correct answer. Dogs, birds, and dolphins are also known for their ability to learn, but humans are generally considered to have the greatest ability to learn due to their advanced cognitive capabilities and complex societies. Therefore, the correct answer is humans as they have the greatest ability to learn compared to the other options.]
\\

\hline\hline
\multicolumn{3}{p{0.95\textwidth}}{
\textbf{Example}: F.4 \par
\textbf{Dataset}: ARC-Challenge \par
\textbf{Question [\textcolor{blue}{Answer}]}:
 Which items would take the least amount of time to decompose in nature?
{[}\textcolor{blue}{cut grass}{]}
} \\
\hline\hline
\textbf{Model} & \textbf{Generated Distractors} & \textbf{Rationale} \\
\hline
Ground Truth                & [paper cups, newspapers, cut down trees]       & --\\
T5(base)                    & [a sandbag, ice cubes, leaves]      & -- \\
T5(multi-task)              & [sand, ice, water]      & -- \\
T5(contrast)             & [grass, mosses, leaves]     & -- \\
T5(beam)                    & [sand, weeds, moss]     & -- \\
T5(clustering)              & [ice, weeds, a tree]    & -- \\
GPT-3(clustering)           & [glass bottles, aluminum cans, styrofoam]   & -- \\
TinyLlama                   & [a dead bird, a mushroom, a tree trunk]     & -- \\
Mistral(k-NN)                & [plastic bags, aluminum cans, glass bottles] & -- \\
GPT-3(random)               & [plastic bottles, aluminum cans, glass bottles]     &  -- \\
GPT-3(k-NN)                  & [plastic bottles, aluminum cans, glass bottles]     & -- \\
GPT-3(COT-random)           & [plastic water bottle, aluminum can, glass bottle]  
&  [Cut grass would take the least amount of time to decompose in nature because it is an organic material and can be broken down by natural processes such as bacteria and fungi. In contrast, plastic water bottles, aluminum cans, and glass bottles are all non-biodegradable materials and would take much longer to decompose, if at all. Therefore, cut grass is the correct answer as it would decompose the fastest in nature.]\\
GPT-3(COT-k-NN)              & [plastic bottles, aluminum cans, glass bottles]
&  [Cut grass would take the least amount of time to decompose in nature because it is an organic material that can easily break down with the help of microorganisms and other decomposers. Plastic bottles, aluminum cans, and glass bottles are all non-biodegradable materials that can take hundreds of years to decompose in nature, making them much slower to break down compared to organic materials like cut grass.] \\
\hline\hline
\end{tabular}
}
\caption{Representative examples 
({\bf F.3}, {\bf F.4}) 
of distractor generation models across selected datasets. Underline indicates cases where the generated distractor matches the correct answer.}
\label{tab:T9}
\end{table*}

\begin{table*}[!t]
\centering
\small
\resizebox{\textwidth}{!}{
\begin{tabular}{p{0.20\textwidth} p{0.40\textwidth} p{0.44\textwidth}}
\hline\hline
\multicolumn{3}{p{0.95\textwidth}}{
\textbf{Example}: F.5 \par
\textbf{Dataset}: MCQ \par
\textbf{Question [\textcolor{blue}{Answer}]}:
The average human body contains 5,830 g of
{[}\textcolor{blue}{blood}{]}
} \\
\hline\hline
\textbf{Model} & \textbf{Generated Distractors} & \textbf{Rationale} \\
\hline
Ground Truth                &  [muscle, bacteria, water]    & --\\
T5(base)                    &  [urine, saliva, lungs]   &  --\\
T5(multi-task)              &  [urine, urine, saliva]   & -- \\
T5(contrast)             &  [urine, fat, urine]   & --\\
T5(beam)                    &  [urine, \underline{blood}, saliva]   & --\\
T5(clustering)              &  [liver, fat, bone] &  --\\
GPT-3(clustering)           &  [water, fat, calcium]  & -- \\
TinyLlama                   &  [sand, water, bones]   & -- \\
Mistral(k-NN)                &  [water, muscle tissue, bone]   & -- \\
GPT-3(random)               &  [water, bone, muscle]   & -- \\
GPT-3(k-NN)                  &  [bone, muscle, fat]  &  -- \\
GPT-3(COT-random)           &  [water, bone, muscle]  
&   [The question is asking for the substance that the average human body contains 5,830 g of. The correct answer is  blood as it is a substance found in the human body and the average amount is specified in the question. The distractors water, bone, and  muscle are also substances found in the human body, but the question specifically mentions the amount of a substance, which is blood in this case.]\\
GPT-3(COT-k-NN)              &[water, fat, muscle]  
& [The average human body contains approximately 7-8\% of its weight as blood, which equates to around 4.5-6 liters of blood in an average adult. Water is a major component of the human body, but the average human body contains around 60\% water, not 5,830 g specifically. Fat and muscle are also major components of the human body, but their average quantities in the body do not match the specific weight mentioned in the question. Therefore, the correct answer is blood.]\\
\hline\hline

\multicolumn{3}{p{0.95\textwidth}}{
\textbf{Example}: F.6 \par
\textbf{Dataset}: SciQ \par
\textbf{Question [\textcolor{blue}{Answer}]}:
A bee will sometimes do a dance to tell other bees in the hive where to find what?
{[}\textcolor{blue}{food}{]}
} \\
\hline\hline
\textbf{Model} & \textbf{Generated Distractors} & \textbf{Rationale} \\
\hline
Ground Truth                & [enemies, water, honey]   & --\\
T5(base)                    & [water, \underline{food}, water]      & -- \\
T5(multi-task)              & [water, shelter, \underline{food}]    & -- \\
T5(contrast)             & [honey, eggs, shelter]    & -- \\
T5(beam)                    & [water, \underline{food}, pollen]     & -- \\
T5(clustering)              & [water, nectar, insects]  & -- \\
GPT-3(clustering)           & [shelter, drones, nectar] & -- \\
TinyLlama                   & [flowers, trees, safety]  & -- \\
Mistral(k-NN)                & [water, nectar, pollen]   & -- \\
GPT-3(random)               & [water, shelter, pollen]  & -- \\
GPT-3(k-NN)                  & [water, shelter, pollen]  & -- \\
GPT-3(COT-random)           & [water, shelter, pollen]  
& [The behavior of a bee doing a dance to communicate with other bees in the hive is known as the waggle dance. The purpose of the waggle dance is to communicate the location of a food source, such as nectar or pollen, to other members of the hive. Therefore, the correct answer is food. The distractors water, shelter, and pollen are not typically communicated through the waggle dance.]\\

GPT-3(COT-k-NN)              & [water, shelter, predators]  &
[Bees communicate the location of food sources to other bees
through a dance known as the waggle dance. The dance conveys information about
the direction and distance of the food source. Therefore, the purpose of the dance
is to tell other bees where to find food. The distractors provided (water, shelter,
predators) do not accurately represent the information communicated through the
waggle dance, as it specifically relates to the location of food sources.] \\
\hline\hline
\end{tabular}
}
\caption{Representative examples ({\bf F.5}, {\bf F.6}) of distractor generation models across selected datasets. Underline indicates cases where the generated distractor matches the correct answer.}
\label{tab:T10}
\end{table*}

\begin{table*}[!t]
\centering
\small
\resizebox{\textwidth}{!}{
\begin{tabular}{p{0.18\textwidth} p{0.40\textwidth} p{0.44\textwidth}}
\hline\hline

\multicolumn{3}{p{0.95\textwidth}}{
\textbf{Example}: F.7 \par
\textbf{Dataset}: MCQL \par
\textbf{Question [\textcolor{blue}{Answer}]}:
Main sources of carbohydrates are
{[}\textcolor{blue}{green plants}{]}
} \\
\hline\hline
\textbf{Model} & \textbf{Generated Distractors} & \textbf{Rationale} \\
\hline
Ground Truth                & [milk, water, proteins]  & --\\
T5(base)                    & [red plants, blue plants and green plants alike, yellow plants]   & -- \\
T5(multi-task)              & [red plants, blue plants and green plants alike, yellow plants]    & -- \\
T5(contrast)             & [animals, plants, fish]    & -- \\
T5(beam)                    & [red plants, white plants, brown plants]     & -- \\
T5(clustering)              & [red plants, green vegetables, starch] & -- \\
GPT-3(clustering)           & [animal products, fast food, water sources] & -- \\
TinyLlama                   & [herbivores, red meat, processed foods]   & -- \\
Mistral(k-NN)                & [animals, fungi, bacteria]  & -- \\
GPT-3(random)               & [meat, dairy products, fish]    &  -- \\
GPT-3(k-NN)                  & [meat, dairy products, seafood]   & -- \\
GPT-3(COT-random)           & [meat, dairy products, fish ]  
& [Green plants are the main sources of carbohydrates, as they produce carbohydrates through the process of photosynthesis. Meat, dairy products, and fish are sources of protein and fat, but they are not the main sources of carbohydrates. Therefore, the correct answer is green plants, as they are the primary source of carbohydrates in the diet.] \\
GPT-3(COT-k-NN)              
& [meat and dairy products, seafood, nuts and seeds]  
&[The main sources of carbohydrates are green plants, such as fruits, vegetables, grains, and legumes. These plant-based foods are rich in carbohydrates, providing energy and essential nutrients for the body. Meat and dairy products are not the main sources of carbohydrates, as they primarily provide protein and fats. Seafood is also not a significant source of carbohydrates. Nuts and seeds contain some carbohydrates, but they are not the main sources compared to green plants.   Therefore, the correct answer is green plants.]\\
\hline\hline

\multicolumn{3}{p{0.95\textwidth}}{
\textbf{Example}: F.8 \par
\textbf{Dataset}: ARC-Easy \par
\textbf{Question [\textcolor{blue}{Answer}]}:
George is studying erosion. He learns how rocks break apart over time. Which of these most likely forms  when a rock weathers and breaks into small pieces?
{[}\textcolor{blue}{sediments}{]}
} \\
\hline\hline
\textbf{Model} & \textbf{Generated Distractors} & \textbf{Rationale} \\
\hline
Ground Truth                & [fossils, crystals, minerals]     & --\\
T5(base)                    & [igneous rocks, sandstones, \underline{sediments}]     & -- \\
T5(multi-task)              & [igneous rocks, sandstones, \underline{sediments}]   & -- \\
T5(contrast)             & [crystals, igneous rocks, metamorphic rocks]   & -- \\
T5(beam)                    & [\underline{sediments}, volcanic eruptions, sand]   & -- \\
T5(clustering)              & [sand, volcanic activity]   & -- \\
GPT-3(clustering)           & [fossils, lava, glaciers]    & -- \\
TinyLlama                   & [igneous, metamorphic, mappable rocks]    & -- \\
Mistral(k-NN)                & [crystals, igneous rocks, metamorphic rocks]  & -- \\
GPT-3(random)               & [fossils, minerals, lava]   &  -- \\
GPT-3(k-NN)                  & [magma, fossils, minerals]   & -- \\
GPT-3(COT-random)           & [minerals, fossils, magma]  
& [Sediments are small pieces of rock that are the result of weathering and erosion. When rocks break apart over time, they form sediments.Minerals are the individual components of rocks and are not formed when rocks break into small pieces. Fossils are the preserved remains or traces of ancient organisms and are not formed directly from the weathering and breaking of rocks. Magma is molten rock beneath the Earth's surface and is not formed from the weathering and breaking of rocks.Therefore, the most likely formation when a rock weathers and breaks into small pieces is sediments.] \\
GPT-3(COT-k-NN)              & [magma, fossils, crystals] 
& [Sediments: When a rock weathers and breaks into small pieces, these small pieces are called sediments. This is the most likely outcome of the weathering and breaking apart of rocks. Magma: Magma is molten rock beneath the Earth's surface and is not directly formed from the weathering and breaking apart of rocks. Fossils: Fossils are the preserved remains or traces of animals, plants, and other organisms from the past, and are not directly formed from the weathering and breaking apart of rocks. Crystals: Crystals are formed through the process of crystallization from a liquid or gas, and are not directly formed from the weathering and breaking apart of rocks. Therefore, the most likely form when a rock weathers and breaks into small pieces is sediments. ] 
\\
\hline\hline
\end{tabular}
}
\caption{Representative examples ({\bf F.7}, {\bf F.8}) of distractor generation models across selected datasets. Underline indicates cases where the generated distractor matches the correct answer.}
\label{tab:T11}
\end{table*}

\begin{table*}[!t]
\centering
\small
\resizebox{\textwidth}{!}{
\begin{tabular}{p{0.18\textwidth} p{0.40\textwidth} p{0.44\textwidth}}
\hline\hline
\multicolumn{3}{p{0.95\textwidth}}{
\textbf{Example}: F.9 \par
\textbf{Dataset}: MedQA \par
\textbf{Question [\textcolor{blue}{Answer}]}:
A 24-year-old pregnant woman at 28 weeks gestation presents to the emergency department with complaints of fever with chills and pain in her knee and ankle joints for the past 2 days. She also complains of headaches and difficulty moving her neck. Further questioning reveals that she had a tick bite on her arm while gardening a few days ago. Past medical history is noncontributory. She takes a multivitamin with iron and folate every day and has been receiving regular prenatal care and the pregnancy is progressing  normally. On examination, an erythematous rash is seen on her right arm, as shown in the accompanying photograph. Her obstetric examination is normal.  Ultrasound of the fetus is reassuring with a normal heartbeat and no gross abnormalities. A specimen is collected to test for Lyme disease. What is the next best step for this patient?
{[}\textcolor{blue}{Amoxicillin}{]}
} \\
\hline\hline
\textbf{Model} & \textbf{Generated Distractors} & \textbf{Rationale} \\
\hline
Ground Truth                & [Ibuprofen, Doxycycline, Tetracycline, Gentamicin]   & --\\
T5(base)                    & [\underline{Amoxicillin}, Vaccination, Lyme disease screening, Immunoglobulin]     & -- \\
T5(multi-task)              &  [Amphotericin B, Vaccination against Lyme disease, Antibiotics,  Vitamin B12]    & -- \\
T5(contrast)             & [Antibiotics, Iron supplementation, Antistreptolysin O titer, Lysosomal trafficking screen]      & -- \\
T5(beam)                    & [Ceftriaxone, Amphotericin B, Azithromycin]    & -- \\
T5(clustering)              & [Ceftriaxone, Amphotericin, Antibiotics]     & -- \\
GPT-3(clustering)           & [Ceftriaxone, Azithromycin, Penicillin G]    & -- \\
TinyLlama                   & [No treatment indicated, Penicillin, Doxycycline]    & -- \\
Mistral(k-NN)                & [Doxycycline, Cefotaxime, Azithromycin]  & -- \\
GPT-3(random)               & [Doxycycline, Ceftriaxone, Acetaminophen, Prednisone]  &  -- \\
GPT-3(k-NN)                  & [Doxycycline, Ceftriaxone, Azithromycin, Penicillin]    & -- \\
GPT-3(COT-random)           & [Doxycycline, Ceftriaxone, Acetaminophen, Ibuprofen]
&  [The patient's symptoms, including fever, joint pain, headaches, and neck stiffness, along with the history of a tick bite and the presence of an erythematous rash at the site of the bite, are highly suggestive of Lyme disease. The next best step for this patient is to start treatment with amoxicillin, which is the first-line antibiotic for the treatment of Lyme disease in pregnant women. Doxycycline is contraindicated in pregnancy as it can affect fetal bone and teeth development. Ceftriaxone is not the first-line treatment for Lyme disease in pregnancy. Acetaminophen and ibuprofen may help with symptom management, but the priority is to start antibiotic treatment for the suspected Lyme disease. In conclusion, the next best step for this patient is to start treatment with amoxicillin for suspected Lyme disease.] \\
GPT-3(COT-k-NN)              & [Doxycycline, Ceftriaxone, Azithromycin, Penicillin]
& [The patient's history of a tick bite, along with the symptoms of fever, joint pain, headaches, and the characteristic erythematous rash, are highly suggestive of Lyme disease. The next best step for this patient would be to start treatment with amoxicillin, which is the recommended antibiotic for treating Lyme disease in pregnant women. Doxycycline is contraindicated in pregnancy as it can affect fetal bone and teeth development. Ceftriaxone is not the first-line treatment for Lyme disease in pregnancy. Azithromycin is not the first-line treatment for Lyme disease, and penicillin is not the preferred antibiotic for treating Lyme disease.]
\\
\hline\hline
\end{tabular}
}
\caption{Representative example ({\bf F.9}) of distractor generation models across selected datasets. Underline indicates cases where the generated distractor matches the correct answer.}
\label{tab:supp_examples}
\end{table*}

\begin{table*}[!t]
\centering
\small
\resizebox{\textwidth}{!}{
\begin{tabular}{p{0.18\textwidth} p{0.40\textwidth} p{0.44\textwidth}}
\hline\hline
\multicolumn{3}{p{0.95\textwidth}}{
\textbf{Example}: F.10 \par
\textbf{Dataset}: MedQA \par
\textbf{Question [\textcolor{blue}{Answer}]}:
The parents of a 14-year-old patient are concerned and have questions about the use of insulin for their son's recently diagnosed type 1 diabetes. The patient has developed an upper respiratory infection while at school. He is coughing and has a runny nose. His temperature is $37.8\,^{\circ}\mathrm{C}\ (100.2\,^{\circ}\mathrm{F})$ and vital signs are within normal limits. Physical examination is unremarkable. Which of the following modifications to his insulin regimen would you recommend to this patient and his parents? 
{[}\textcolor{blue}{Increase the frequency of blood glucose checks}{]}
} \\
\hline\hline
\textbf{Model} & \textbf{Generated Distractors} & \textbf{Rationale} \\
\hline
Ground Truth                & [Reduce the insulin dose, Continue same regimen, Hold insulin until the patient gets better, Increase the insulin dose to double]           & --\\
T5(base)                    & [\underline{Increase the frequency of blood glucose checks}, Increase dietary intake of glucagon, Obtain a blood sugar level]           & -- \\
T5(multi-task)              & [\underline{Increase the frequency of blood glucose checks}, Increase his daily insulin intake, Add a small amount of glycemic control]           & -- \\
T5(contrast)             & [Increase the frequency of insulin checks, Add a small amount of water per day to his insulin regimen, Switch to glucagon]           & -- \\
T5(beam)                    & [\underline{Increase the frequency of blood glucose checks}, Increase the frequency of insulin doses, Increase the duration of insulin therapy]           & -- \\
T5(clustering)              &  [Increase the frequency of insulin intake, Increase the amount of insulin he takes]          & -- \\
GPT-3(clustering)           & [Use insulin only when the patient is physically active]           & -- \\
TinyLlama                   &  [\underline{Increase the frequency of blood glucose checks}, Administer insulin every 3 hours, Administer insulin every hour and provide pre-meal]          & -- \\
Mistral(k-NN)                & [Decrease the dose of insulin, Stop insulin therapy temporarily, Continue insulin therapy as prescribed without modification] & -- \\
GPT-3(random)               & [Increase the dose of long-acting insulin, Switch to oral hypoglycemic agents, Decrease the frequency of insulin injections, Increase the dose of short-acting insulin]           &  -- \\
GPT-3(k-NN)                  & [Increase the insulin dosage, Decrease the insulin dosage, Switch to a different type of insulin, Stop insulin therapy temporarily]           & -- \\
GPT-3(COT-random)           & [Increase the dose of insulin, Switch to a different type of insulin, Decrease the frequency of blood glucose checks, Stop insulin temporarily]
&  [The patient has developed an upper respiratory infection, which can affect blood glucose levels and insulin requirements. It is important to increase the frequency of blood glucose checks to monitor for any changes in blood sugar levels due to the illness. Increasing the dose of insulin without monitoring blood glucose levels could lead to hypoglycemia. Switching to a different type of insulin may not be necessary if the patient's blood glucose levels can be managed with increased monitoring. Decreasing the frequency of blood glucose checks would not be appropriate, especially during an illness. Stopping insulin temporarily is not recommended, as it could lead to uncontrolled hyperglycemia.Therefore, the most appropriate modification to the insulin regimen would be to increase the frequency of blood glucose checks to closely monitor the patient's blood sugar levels during the illness.] \\
GPT-3(COT-k-NN)              & [Increase the insulin dosage, Switch to a different type of insulin, Decrease the frequency of blood glucose checks, Stop insulin therapy temporarily ]
& [The patient has developed an upper respiratory infection, which can affect blood sugar levels and insulin requirements. It is important to increase the frequency of blood glucose checks to closely monitor the patient's blood sugar levels during the illness. Increasing the insulin dosage without proper monitoring can lead to hypoglycemia, so it is not the most appropriate recommendation in this situation. Switching to a different type of insulin may not be necessary at this time, as the focus should be on closely monitoring blood sugar levels. Decreasing the frequency of blood glucose checks is not recommended, especially during an illness when blood sugar levels can fluctuate. Therefore, the most appropriate modification to the insulin regimen would be to increase the frequency of blood glucose checks to closely monitor the patient's blood sugar levels during the upper respiratory infection.]
\\
\hline\hline
\end{tabular}
}
\caption{Representative example ({\bf F.10}) of distractor generation models across selected datasets. Underline indicates cases where the generated distractor matches the correct answer.}
\label{tab:T12}
\end{table*}

\end{document}